\documentclass[10pt,twocolumn,letterpaper]{article}

\usepackage{cvpr}              

\usepackage{graphicx}
\usepackage{amsmath}
\usepackage{amssymb}
\usepackage{booktabs}

\allowdisplaybreaks[4]
\usepackage{algorithm}
\usepackage{algorithmic}
\usepackage{newfloat}
\usepackage{listings}
\floatstyle{ruled}
\newfloat{listing}{tb}{lst}{}
\floatname{listing}{Listing}

\usepackage{breqn}
\usepackage{multirow}
\usepackage{makecell}
\usepackage{verbatim}

\usepackage{algorithm}
\usepackage{listings}

\newtheorem{theorem}{Theorem}[section]

\newtheorem{lemma}[theorem]{Lemma}
\newtheorem{proof}{Proof}[section]

\newcommand{\twonorm}[1]{\bigg{\lVert}{#1}\bigg{\rVert} _2}
\newcommand{\lnorm}[1]{\frac{#1}{\left\lVert{#1}\right\rVert _2}}

\usepackage[dvipsnames]{xcolor}
\usepackage[breaklinks,colorlinks,citecolor = ForestGreen]{hyperref}

\usepackage[capitalize]{cleveref}
\crefname{section}{Sec.}{Secs.}
\Crefname{section}{Section}{Sections}
\Crefname{table}{Table}{Tables}
\crefname{table}{Tab.}{Tabs.}

\def\cvprPaperID{1617} 
\def\confName{CVPR}
\def\confYear{2022}

\begin{document}

\title{Consistency Regularization for Deep Face Anti-Spoofing}

\author{
Zezheng Wang$^{1,}$\thanks{denotes equal contribution.}~~~~Zitong Yu$^{2,*}$~~~~Xun Wang$^{1}$~~~~Yunxiao Qin$^{3}$~~~~Jiahong Li$^{1,}$\thanks{denotes corresponding author.}~~~~Chenxu Zhao$^{5, \dagger}$\\
Zhen Lei$^4$~~~~Xin Liu$^{6}$~~~~Size Li$^1$~~~~Zhongyuan Wang$^1$\\[1mm]
\normalsize{
$^1$Kuaishou Technology~~~~
$^2$University of Oulu~~~~
$^3$Communication University of China~~~
$^4$CBSR\&NLPR, CASIA
}\\
\normalsize{
$^5$Academy of Sciences, Mininglamp Technology~~~~
$^6$Lappeenranta-Lahti University of Technology
}
}

\maketitle

\begin{abstract}
   Face anti-spoofing (FAS) plays a crucial role in securing face recognition systems. Empirically, given an image, a model with more consistent output on different views of this image usually performs better, as shown in Fig.~\ref{fig:fig1_inro}. Motivated by this exciting observation, we conjecture that encouraging feature consistency of different views may be a promising way to boost FAS models. In this paper, we explore this way thoroughly by enhancing both \textbf{E}mbedding-level and \textbf{P}rediction-level \textbf{C}onsistency \textbf{R}egularization (EPCR) in FAS. Specifically, at the embedding-level, we design a \emph{dense similarity loss} to maximize the similarities between all positions of two intermediate feature maps in a self-supervised fashion; while at the prediction-level, we optimize the mean square error between the predictions of two views. Notably, our EPCR is free of annotations and can directly integrate into semi-supervised learning schemes. Considering different application scenarios, we further design five diverse semi-supervised protocols to measure semi-supervised FAS techniques. We conduct extensive experiments to show that EPCR can significantly improve the performance of several supervised and semi-supervised tasks on benchmark datasets. The codes and protocols will be released at  \url{https://github.com/clks-wzz/EPCR}.
\end{abstract}
\vspace{-1.0em}

\section{Introduction}
\label{sec:intro}

\begin{figure}[ht]
  \centering
  \begin{subfigure}[b]{0.23\textwidth}
    \includegraphics[width=1.0\linewidth]{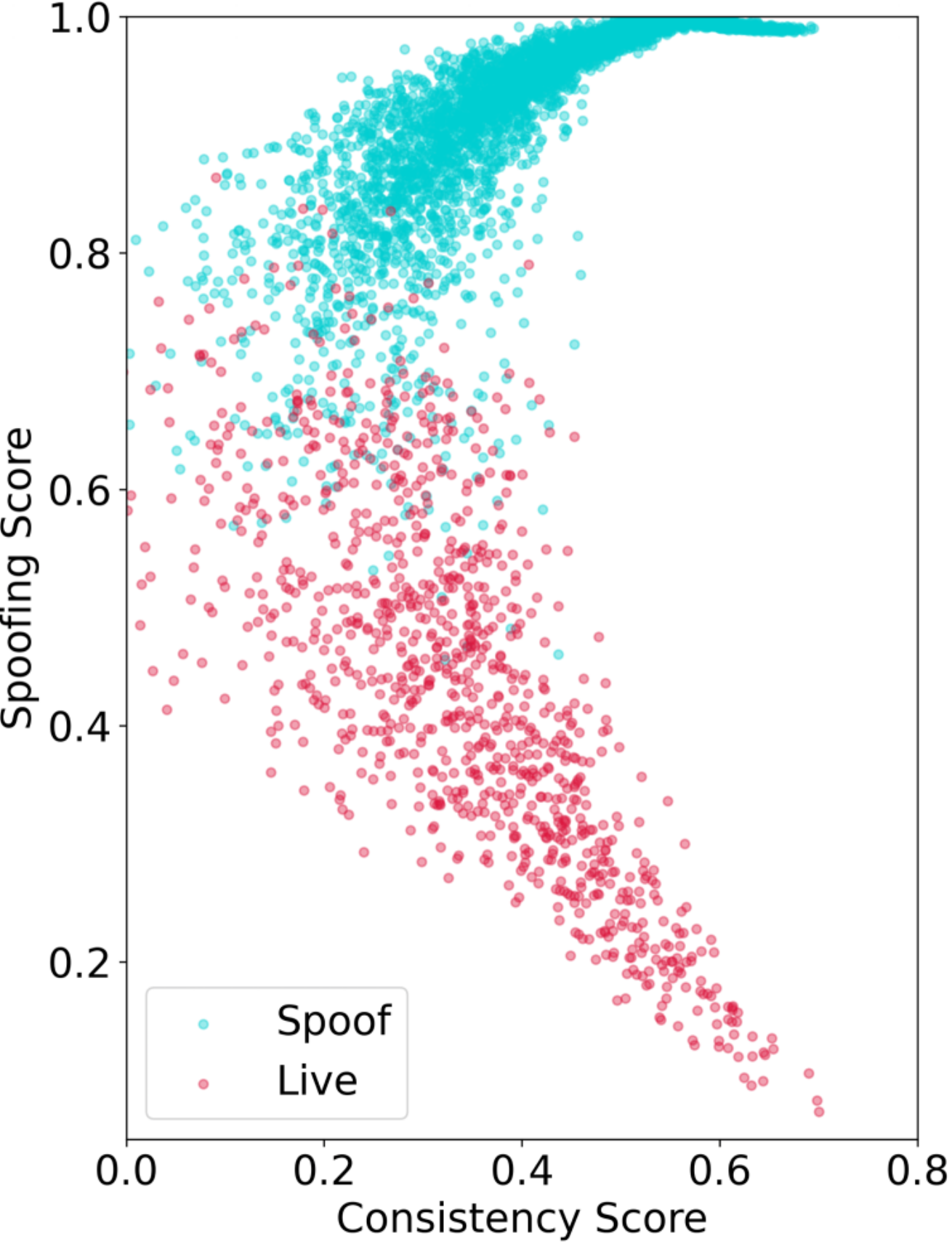}
    \caption{Consistency of samples}
    \label{fig:subfig:a} 
  \end{subfigure}
  \begin{subfigure}[b]{0.23\textwidth}
    \includegraphics[width=1.0\linewidth]{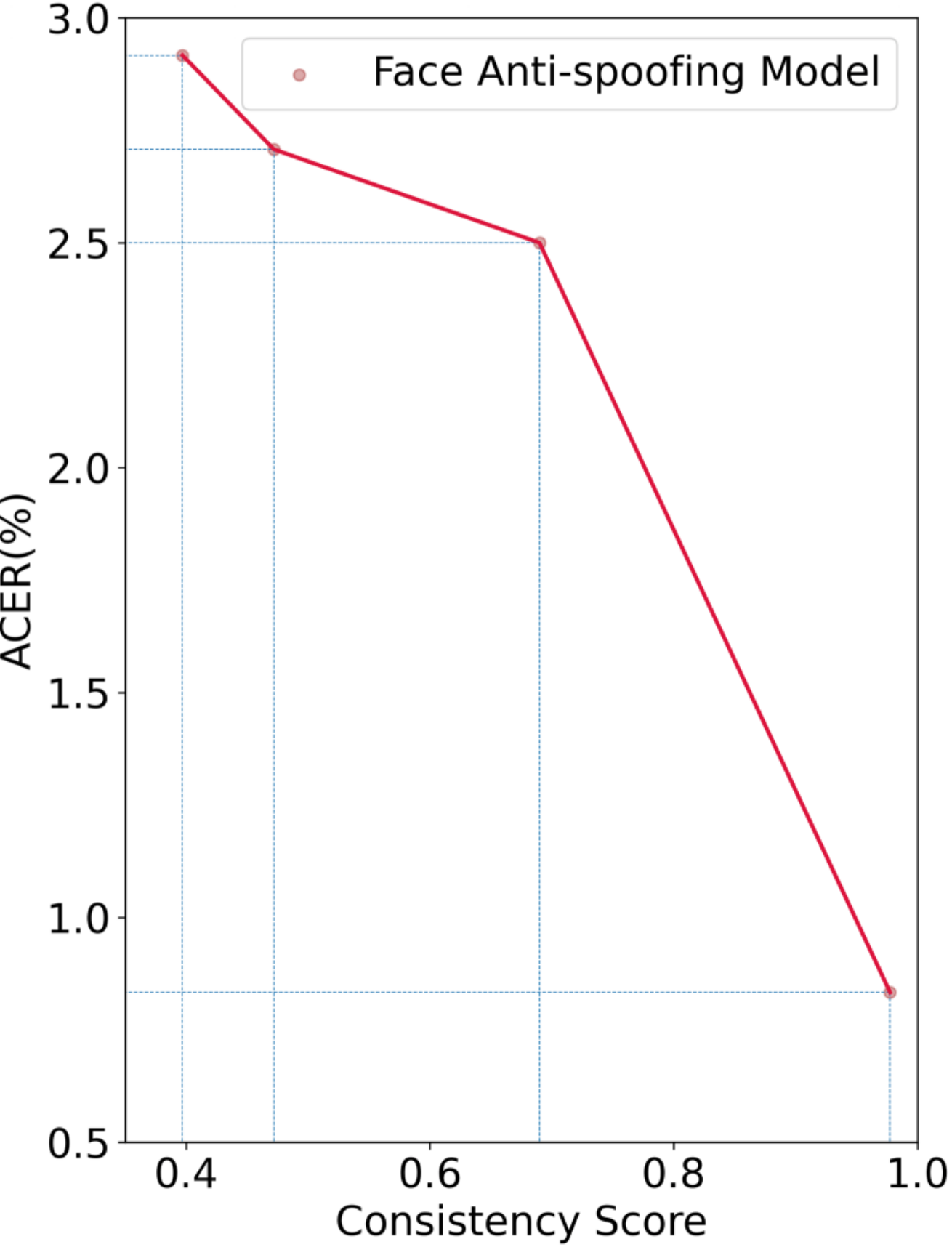}
    \caption{Consistency of models}
    \label{fig:subfig:b} 
  \end{subfigure}
  \caption{\textbf{Consistency for samples and models} on Protocol 1 of OULU-NPU~\cite{Boulkenafet2017OULU}. Spoofing score is the probability that one sample is predicted as an attack. Lower ACER means better model. Consistency score is the similarity between features of two augmented views. {\bf (a)}: samples with higher consistency tends to have smaller errors. {\bf (b)}: models with higher consistency scores perform better.}
  \label{fig:fig1_inro} 
  \vspace{-0.5em}
\end{figure}

Nowadays, face recognition~\cite{deng2019arcface} has been an indispensable component of AI systems and is successfully used in various applications, such as mobile payments and guard entrances. However, existing face recognition systems are still vulnerable to diverse presentation attacks (PAs), including print attack, replay attack, and 3D mask attack, which lead to a serious security risk. To address this issue, face anti-spoofing (FAS)~\cite{ming2020survey,yu2021deep} techniques are proposed and have attracted considerable attention in the industry and research communities.

Recently, to discriminate the live and spoof faces, methods based on hand-crafted descriptors~\cite{Pereira2012LBP,Patel2016Secure,Boulkenafet2017Face_SURF} and those based on deep learning~\cite{Lucena2017Transfer,Liu2018Learning,yu2020fas2,george2021cross,george2021effectiveness,qin2021meta} have been proposed. Specifically, 
hand-crafted descriptors are the spoofing clues (\eg, the loss of skin details, color distortion, moir$\rm\acute{e}$ pattern, motion pattern, and spoofing artifacts,~\etc), that can distinguish whether a sample is a live face. While, deep learning FAS methods utilize deep networks ~\cite{yu2020searching} and annotated samples~\cite{Liu2018Learning,wang2020deep} to learn discriminative features automatically. Though the effectiveness of the above methods, FAS faces a critical challenge of domain gap: performance degradation on unseen domains. \cite{li2018domain, shao2019multi,jia2020single} align the distributions among multiple source domains to learn more generalized features.

For human perception and cognition~\cite{garling1989environmental}, when an object changes slightly, it is still regarded as the same object~\cite{tarvainen2017mean}. Inspired by this idea, we begin to explore feature (including intermediate embedding and final prediction) consistency in FAS. In the real world, the live and spoof faces from diverse scenarios of~\eg~different illuminations, sensors, and spoof types, can result in different features. With experiments on Protocol-1 of OULU-NPU~\cite{Boulkenafet2017OULU} using CDCN~\cite{yu2020searching} in Fig.~\ref{fig:subfig:a}, we observe an interesting phenomenon that samples with higher consistency under different augmentations usually have lower prediction errors. To move one step further, we conjecture that ensuring feature invariance under different input perturbations may benefit a model's performance.

Nevertheless, in our experimental observation (CDCN vs. SimSiam-CDCN in Tab.~\ref{tab:OULU}), directly introducing traditional consistency regularization method even hinders the generalization of FAS models. Therefore, it is still challenging to exploit an effective formulation on consistency regularization to improve the performance of FAS.
Motivated by the above inspiration, we propose a novel FAS framework through regularizing consistency on embedding and prediction levels (EPCR for short) to ensure effective feature consistency and provide training signals even without annotations.  Particularly, at the embedding-level, we design a \emph{dense similarity objective} to optimize the dense relationships between all positions of intermediate feature maps in a self-supervised fashion. This objective requires close features at all positions of different views of a given face image, since different regions from different views of a facial image usually share the same live/spoof pattern. At the prediction level, we use a mean square error objective to regularize the output consistency between two augmentations. 
As shown in Fig.~\ref{fig:subfig:b}, benefits from the two regularizations, EPCR has higher feature consistency and better performance than with models without such regularization.

Note that our EPCR can be viewed as a self-supervised auxiliary task like the pretext tasks in unsupervised and semi-supervised learning~\cite{sajjadi2016regularization, laine2016temporal, tarvainen2017mean, xie2019unsupervised,  deng2009imagenet, scudder1965probability, lee2013pseudo, sohn2020fixmatch}. Thus, it can be adopted not only in fully-supervised FAS but also in semi-supervised FAS, which is another advantage of the proposed EPCR. 
As a large amount of unlabelled face data can be easily collected (\eg, from access control system), semi-supervised FAS that learns from both limited labeled and extra unlabeled data may be a promising direction to more generalized and more stable FAS models.
To verify this possibility, we establish several new semi-supervised FAS benchmarks. 
The main contributions of this work are summarized below.
\vspace{-0.6em}
\begin{itemize}
\setlength\itemsep{-0.1em}
\item We discover the interesting relationship between the output consistency of a deep FAS model and the model's performance on presentation attack detection.
\item We propose a novel embedding-level and prediction-level consistency regularization (EPCR) framework to ensure the FAS model's feature consistency and to provide self-supervised learning signals. 
\item We build five diverse semi-supervised FAS benchmarks, covering various real-world scenarios.
\item We test EPCR in both fully-supervised and semi-supervised FAS scenarios. With extensive experiments on widely used benchmarks, we demonstrate the state-of-the-art performance of EPCR.
\end{itemize}

\section{Related Work}
\vspace{-0.5em}
\label{subsec:realte_fas}
\textbf{Face Anti-Spoofing}.  Traditional face anti-spoofing (FAS) methods usually extract hand-crafted descriptors,~\eg, LBP~\cite{boulkenafet2015face} and HOG~\cite{Komulainen2014Context}, from the facial images to capture the spoofing patterns. Some deep learning based methods treat the FAS problem as a binary classification task~\cite{Li2017An}, and utilize binary cross-entropy loss to optimize the model. Motivated by physical discrepancy between live and spoof faces, dense pixel-wise supervisions~\cite{yu2020revisiting},~\eg, pseudo depth map~\cite{Liu2018Learning,yu2020searching,wang2020deep,yu2021dual}, reflection map~\cite{zhang2020celeba,yu2020face}, texture map~\cite{zhang2020face} and binary map~\cite{george2019deep,liu2019deep} are introduced for learning intrinsic features. 
FAS can be also formulated as a domain adaptation/generalization problem in ~\cite{li2018domain, jia2020single,  wang2020cross}, where disentangled learning~\cite{zhang2020face,liu2020disentangling}, adversarial learning~\cite{shao2019multi}, and meta learning~\cite{shao2020regularized,chen2021generalizable,wang2021self,liu2021dual,liu2021adaptive} are adopt to improve the model's generalization capacity on unseen scenarios. Besides, several anomaly detection~\cite{li2020unseen,nikisins2018effectiveness,baweja2020anomaly,fatemifar2021client}, zero-shot~\cite{liu2019deep,Qin2020MetaFas} and continuous learning~\cite{rostami2021detection,perez2020learning} approaches are proposed for unknown PAs detection. Though the generalization capacity has been improved by those techniques, they are still restricted to labeled data in source domains.
Recently, several works~\cite{jia2021unified, 2021Progressive} start introducing the unlabeled data into FAS training, while semi-supervised benchmarks with comprehensive scenarios (\eg, cross domains and attack types) have not been built yet.

\vspace{0.3em}
\textbf{Consistency Regularization}. 
Consistency regularization aims to ensure the model outputs be stable to the small input perturbations, and is widely used in self-supervised learning~\cite{chen2020simple, grill2020bootstrap} and semi-supervised learning~\cite{sajjadi2016regularization, laine2016temporal, tarvainen2017mean, xie2019unsupervised}. 
Self-supervised methods~\cite{chen2020simple, grill2020bootstrap} regard two different augmentations of an image as a positive pair and use the contrastive loss to pull the positive pair closer. This feature consistency regularization is also leveraged in semi-supervised learning ~\cite{tarvainen2017mean,sohn2020fixmatch} to improve the performance.
Specifically, these semi-supervised methods optimize the similarity between two outputs of different augmented views of one unlabeled image and the standard supervised objective on labeled images. Despite success in general self-supervised and semi-supervised tasks, consistency regularization methods are still unexplored in FAS with serious domain shift and unknown attack issues.

\vspace{-0.3em}
\section{Methodology}

As illustrated in Fig.~\ref{fig:overall_framework}, our proposed \textbf{E}mbedding-level and \textbf{P}rediction-level \textbf{C}onsistency \textbf{R}egularization (\textbf{EPCR}) for face anti-spoofing method contains a basic self-supervised architecture that two augmented views of one facial image can learn representations from each other. 
Embedding-level consistency regularization is implemented with a dense feature generator and dense similarity loss, which assists the model in learning more discriminative representation. In order to further improve the model's stability, prediction-level consistency regularization is also introduced via a dense classifier and Mean Square Error loss between the prediction of Siamese networks.

\subsection{Basic Architecture}
Inspired by the idea of classical unsupervised/semi-supervised methods~\cite{ laine2016temporal, tarvainen2017mean, xie2019unsupervised, he2020momentum, grill2020bootstrap, chen2020exploring}, we adopt a Siamese-like network as the basic architecture. 
The overall framework is shown in Fig.~\ref{fig:overall_framework}. We augment the face image input $x$ into two views ($x_1$ and $x_2$)  and feed them into the dense encoder $f$, composed of a backbone $b$ and a projector $p$, which shares parameters on both sides. CDCN~\cite{yu2020searching} is utilized as the backbone due to its intrinsic feature representation capacity. To facilitate the convergence acceleration with less GPU memory, we plug a max-pooling layer after the first convolution block for downsampling. The feature maps with $s \times s$ spatial resolution from the encoder are represented by $\mathbf{F_1}$ and $\mathbf{F_2}$. In our EPCR framework, Embedding-level consistency regularization with dense similarity loss and prediction-level with MSE smooth loss are the most essential components while the Siamese structure~\cite{chen2020exploring} can be flexibly replaced by alternatives (\eg, SimCLR~\cite{chen2020simple} and BYOL~\cite{grill2020bootstrap}).

\subsection{Embedding-level Consistency Regularization}

Existing contrastive learning based methods either optimize the cosine similarity between the global vectors \cite{grill2020bootstrap,chen2020exploring,wang2020cross}, or learn a consistent representation for a local field \cite{wang2020dense}. 
However, considering the most types of presentation attacks (i.e., print attack, replay attack, and 3D mask) in FAS, each local patch in the facial region  can share the same spoof clues (e.g., material and quality).
Therefore, we can reasonably assume that the vector at each point of $\mathbf{H_1}$ contains the maximal similarity to that of $\mathbf{F_2}$. This is different from the previous contrastive methods.

Motivated by this, we design a \textbf{Embedd.-Consis.} module with a {\it dense predictor h} and a {\it dense  similarity loss}, respectively. 
Compared with the predictor in SimSiam~\cite{chen2020simple}, our dense predictor removes the global pooling layer and replaces the MLP with two $1 \times 1$ convolution layers. Before introducing our new loss function, we first define the dense similarity between two 
flattened tensors $\mathbf{H}, \mathbf{F} \in \mathcal{R}^{s^2\times d}$ as
\vspace{-0.1em}
\begin{equation}
\mathcal{DS}(\mathbf{H}, \mathbf{F}) = \sum_{i} \sum_{j} \lnorm{\mathbf{H}^{i}}{\cdot}\lnorm{\mathbf{F}^{j}},
\label{eq:dense_sim}
\end{equation}
where $\mathbf{H}^{i}$ denotes the $i$-th row of $\mathbf{H}$. 
Noting that the proposed dense similarity in Eq.~\ref{eq:dense_sim} is different from the dense correspondence cross views in \cite{wang2020dense}
, whose correspondence is obtained by applying an argmax operation to the similarity matrix along the last dimension of two features from the backbone.
With the definiton, we can derive the following lemma and proof:
\begin{lemma}
\begin{equation}
\mathcal{DS}(\mathbf{H}, \mathbf{F})  = \sqrt{\mathcal{DS}(\mathbf{H}, \mathbf{H}) \mathcal{DS}(\mathbf{F}, \mathbf{F})} \; cos<\Bar{\mathbf{H}},\Bar{\mathbf{F}}>,
\label{eq:dense_sim_v2}
\end{equation}
where $\Bar{\mathbf{H}}$ (or $\Bar{\mathbf{F}}$) is the center of all normalized row vectors in $\mathbf{H}$ (or $\mathbf{F}$), which can be denoted as $\Bar{\mathbf{H}} = \frac{1}{s^2}\sum_{i}\lnorm{\mathbf{H}^{i}}$.
\end{lemma}
\begin{proof}
\begin{scriptsize}
\begin{align*}
\begin{split}
& \mathcal{DS}(\mathbf{H}, \mathbf{F})  \\ 
& = \sum_{i} \sum_{j} \lnorm{\mathbf{H}^{i}}{\cdot}\lnorm{\mathbf{F}^{j}} \\
& = \sum_{i}\lnorm{\mathbf{H}^{i}} {\cdot} \sum_{j}\lnorm{\mathbf{F}^{j}} \\
& = \twonorm{\sum_{i}\lnorm{\mathbf{H}^{i}}} \twonorm{\sum_{j}\lnorm{\mathbf{F}^{j}}} cos<\sum_{i}\lnorm{\mathbf{H}^{i}}, \sum_{j}\lnorm{\mathbf{F}^{j}}> \\
& =  \twonorm{\sum_{i}\lnorm{\mathbf{H}^{i}}} \twonorm{\sum_{j}\lnorm{\mathbf{F}^{j}}} cos<\Bar{\mathbf{H}},\Bar{\mathbf{F}}> \\ 
& = \sqrt{\sum_{i}\lnorm{\mathbf{H}^{i}} \cdot \sum_{i}\lnorm{\mathbf{H}^{i}}}  \sqrt{\sum_{j}\lnorm{\mathbf{F}^{j}} \cdot \sum_{j}\lnorm{\mathbf{F}^{j}}} cos<\Bar{\mathbf{H}},\Bar{\mathbf{F}}> \\
& = \sqrt{\mathcal{DS}(\mathbf{H}, \mathbf{H})} \sqrt{\mathcal{DS}(\mathbf{F}, \mathbf{F})} cos<\Bar{\mathbf{H}},\Bar{\mathbf{F}}>.
\end{split}
\end{align*}
\end{scriptsize}
\label{eq:dense_sim_v2_proof}
\end{proof}
\vspace{-1.0em}
{\it The above lemma has indicated that our dense similarity not only reflects the global relationship between the two tensors by $ cos<\Bar{\mathbf{H}},\Bar{\mathbf{F}}>$, but also considers the intra-tensor consistency in $\mathcal{DS}(\mathbf{H}, \mathbf{H})$ and $\mathcal{DS}(\mathbf{F}, \mathbf{F})$}.

We compute our dense similarity loss at embedding-level as below:
\begin{equation}
\begin{aligned}
\mathcal{L}_{embedd} = -\frac{1}{2}(\mathcal{DS}(\mathbf{H}_1, stopgrad(\mathbf{F}_2)) 
\\ + \mathcal{DS}(\mathbf{H}_2, stopgrad(\mathbf{F}_1))),
\label{eq:loss_sym1}
\end{aligned}
\end{equation} 
where $\mathbf{F}_i, \mathbf{H}_i$ are the embedding and the output of the predictor from $i$-th view ($i=1,2$),  respectively. 
$stopgrad$ is the stop-gradient operation introduced in ~\cite{chen2020exploring,grill2020bootstrap}.

\subsection{Prediction-level Consistency Regularization}

\begin{figure*}[!htb]
  \centering
  \includegraphics[width=0.98\textwidth]{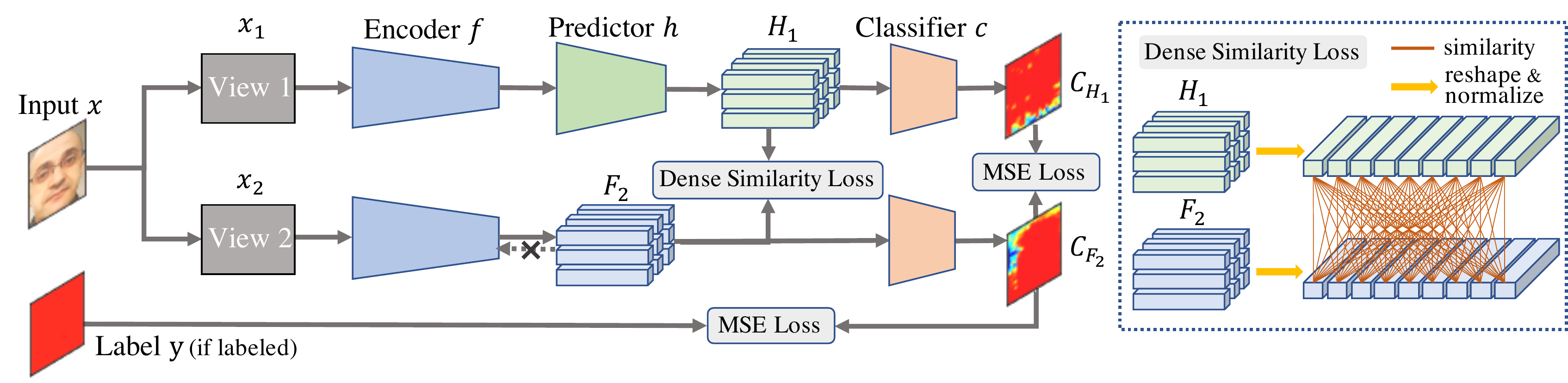}
  \vspace{-8pt}
  \caption{
    \textbf{The framework of EPCR}. For both labeled and unlabeled data, two augmented views ($x_1$ and $x_2$) of the same face image $x$ are processed by the same dense encoder $f$ (a CDCN~\cite{yu2020searching} backbone $b$ and a dense projector $p$). Dense similarity loss between the features ($\mathbf{H_1}$ and $\mathbf{F_2}$) from the dense predictor $h$ of one view and dense encoder $f$ of the other view is constructed for the embedding-level consistency regularization in a stop-gradient manner~\cite{chen2020exploring}. And then the output maps ($\mathbf{C_{H_1}}$ and $\mathbf{C_{F_2}}$) of two views via the same dense classifier $c$ are supervised by mean square error (MSE) loss, which can be seen as output-level consistency regularization. The labeled data is also under the pixel-wise supervision between $\mathbf{C_{F_2}}$ and label $y$. Note that $x_1$ and $x_2$ will be swapped and forward again for symmetrized training. 
  }
  \vspace{-2pt}
  \label{fig:overall_framework}
\end{figure*}
\vspace{-2pt}

The above-mentioned Embedd.-Consis. regularizes the embedding-level consistency between the two generated dense feature maps from the Siamese network. However, there is one concern that the classifier may be out of touch when 
the classifiers of two sides are trained separately. In order to supervise the model more explicitly and stably while further keep the predicted output maps be consistent, we utilize a {\it dense classifier $c$} that shares parameters on both sides, and introduce the Mean Square Error loss between the two outputs, which holds on the prediction-level consistency regularization. 

Dense classifier $c$ is implemented by a $1 \times 1$ convolution layer. And $\mathbf{C_{H_1}}$,  and $\mathbf{C_{F_2}}$ with a $s \times s$ spatial resolution denote the output maps from $H_1$, and $F_2$, respectively. 
Subsequently, the regularization is provided via the MSE loss to supervise the two outputs:
\begin{equation}
\begin{split}
\text{MSE}(\mathbf{C_{H_1}}, \mathbf{C_{F_2}}) = & ||\mathbf{C_{H_1}} - \mathbf{C_{F_2}}||_{2}^{2}.
\label{eq:euclidean_distance_loss}
\end{split}
\end{equation}
At the same time, symmetrized prediction-level consistency regularization (\textbf{Pred.-Consis.}) loss is summarized by:
\begin{equation}
\begin{aligned}
\mathcal{L}_{pred} = \frac{1}{2}\text{MSE}(\mathbf{C_{H_1}}, \mathbf{C_{F_2}}) + \frac{1}{2}\text{MSE}(\mathbf{C_{H_2}}, \mathbf{C_{F_1}}).
\label{eq:loss_sym2}
\end{aligned}
\end{equation} 
Both the embedding-level and prediction-level consistency regularizations can efficiently improve the model's generalization capacity.

\subsection{Overall Loss}
Pixel-wise supervision has been widely used in many face anti-spoofing methods~\cite{Liu2018Learning, yu2020revisiting, yu2020searching, wang2020deep}. In order to make the model learn more discriminative features, we leverage pixel-wise labels in the standard supervision part. Given a scalar binary label $y$ (live/spoof), the binary map label $\mathbf{Y}$ could be obtained via directly expanding $y$. Labels for the spoof are presented as all ones, while those for the live are presented as all zeros. Afterwards, $\mathbf{C_{F_1}}$ and $\mathbf{C_{F_2}}$ are supervised by the square map $\mathbf{Y}$ via MSE loss:
\begin{equation}
\begin{aligned}
\mathcal{L}_{supervised} = \frac{1}{2}\text{MSE}(\mathbf{C_{F_1}}, \mathbf{Y}) + \frac{1}{2}\text{MSE}(\mathbf{C_{F_2}}, \mathbf{Y}).
\label{eq:loss_supervised}
\end{aligned}
\end{equation} 

The overall loss $\mathcal{L}_{overall}$ in the training processing can be formulated as:
\begin{equation}
\begin{aligned}
\mathcal{L}_{overall} = \mathcal{L}_{supervised} + \mathcal{L}_{embedd} + \alpha\mathcal{L}_{pred},
\label{eq:loss_overall}
\end{aligned}
\end{equation} 
where $\alpha$ is the hyperparameter for loss trade-offs, which is empirically set to 0.1. Specially, the $\mathcal{L}_{supervised}$ is omitted for unlabeled samples in semi-supervised learning. 

\section{EPCR for Fully-supervised and Semi-supervised Face Anti-spoofing}
\begin{figure}[!htbp]
\vspace{-1.6em}
\centering
  \includegraphics[width=0.49\textwidth]{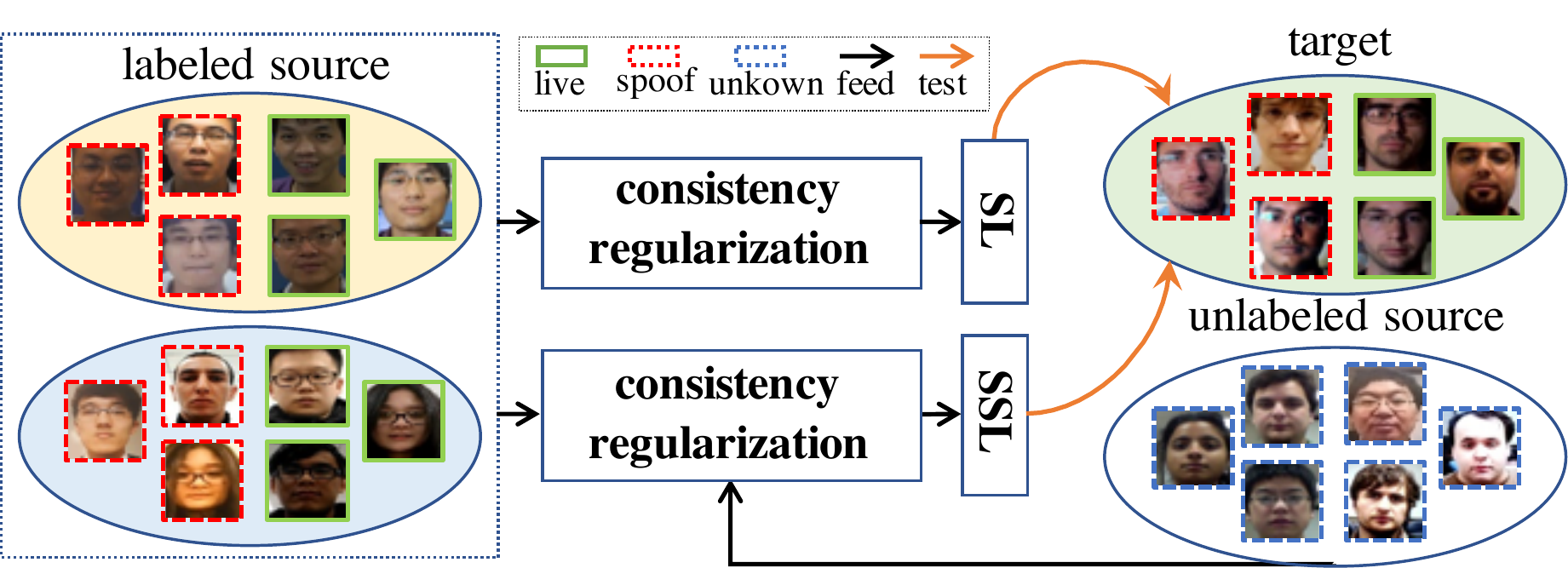}
  \caption{
    The colored oval regions denote face images within different domains. Our proposed EPCR is versatile for both fully-supervised (SL) and semi-supervised (SSL) face anti-spoofing.
  }
  \label{fig:sl_ssl}
   \vspace{-0.8em}
\end{figure}

\subsection{EPCR for Fully-supervised Face Anti-spoofing}

Under the situations with sufficient labeled live and spoof data, it is straightforward to assemble the proposed EPCR on fully-supervised FAS via optimizing Eq.~\ref{eq:loss_overall}. In this way, EPCR benefits the intrinsic live/spoof clues mining via dense regularization on embedding and prediction levels, and alleviates the overfitting issue caused by traditional weak binary supervision.

\noindent \textbf{Fully-supervised learning protocols.} \quad  In our experiments, four public datasets OULU-NPU~\cite{Boulkenafet2017OULU} (denoted as O), CASIA-MFSD~\cite{Zhang2012A} (denoted as C), Replay-Attack~\cite{ReplayAttack} (denoted as I), MSU-MFSD~\cite{wen2015face} (denoted as M) are used for performance verification. In terms of evaluation protocols, we conduct intra-dataset testing on OULU-NPU~\cite{Boulkenafet2017OULU} dataset. Moreover, we conduct domain generalization (DG)~\cite{shao2019multi} cross-dataset testing on all four datasets with a leaving-one-dataset-out protocol. 
More details about these four datasets  and results on DG are shown in the \textsl{Appendix}.

\subsection{EPCR for Semi-supervised Face Anti-spoofing}
\label{sec:protocol}
\noindent \textbf{The insight on semi-supervised FAS.} \quad In real-world FAS applications, we usually encounter a problem that there may be a large margin between the source domain (training set) and the target domain (testing set). For example, the model trained with an indoor database is hard to generalize on the outdoor data due to the variance of outside environments. It is expensive to collect a complete outdoor dataset consisting of both live and spoof faces. Fortunately, there are usually a large amount of face data without labels we can leverage. Therefore, how to make full use of these unlabeled data is a promising problem. In this paper, we also devote ourselves to utilizing the proposed EPCR to address this issue.

\vspace{0.4em}

\noindent \textbf{The feasibility of EPCR on semi-supervised FAS.} \quad Our proposed EPCR could fully exploit the unlabeled data and optimize Eq.~\ref{eq:loss_sym1} and Eq.~\ref{eq:loss_sym2} in a self-supervised manner. Combined with standard supervised learning with partially labeled data, EPCR can be carried out in a semi-supervised FAS task. Introducing the unlabeled data with wider distributions could effectively improve the generalization of the FAS model on the target domain. 
Due to the lack of diverse benchmarks on semi-supervised FAS, we propose five new benchmark protocols to verify the performance of the semi-supervised FAS methods. The proposed \textbf{Semi-supervised learning protocols} are demonstrated as follows:

\vspace{0.4em}
\emph{SSL-Protocol-1}: We follow the original data from standard Protocol 1 of OULU-NPU, in which we set parts of training data as labeled and the other as unlabeled. In this way, the domain of the unlabeled training data have shifted distribution with the testing data from the target domain. 

\vspace{0.4em}
\emph{SSL-Protocol-2}: Here we use extra training data from the training set other than the original ones on Protocol 1 of OULU-NPU, e.g., we use the extra 50\% and 100\% live faces as unlabeled data for semi-supervised learning. This protocol explores the efficacy of the unlabeled data, which shares similar domain information as the target data.

\vspace{0.4em}
\emph{SSL-Protocol-3}: Considering that the data in OULU-NPU is collected by researchers with particular behaviors, we try to conduct more challenging semi-supervised tasks on the domain generalization (DG)~\cite{shao2019multi} cross-dataset evaluation. To be specific, we take one of the three training sets from standard DG protocols as the unlabeled set and the other two as the labeled sets, which can be named as leave-one-unlabeled manner. Unlike \emph{SSL-Protocol-2}, the extra unlabeled training data contains very different domain clues from both the labeled training set and the testing set.

\vspace{0.4em}
\emph{SSL-Protocol-4}: Parts of the data in each training set of standard DG protocols is called out as labeled data and the remaining ones are used as unlabeled data. In this case, compared with the standard DG protocols, there is almost no shortage of domain information, but the amount of labeled data is relatively small.

\vspace{0.4em}
\emph{SSL-Protocol-5}: In this attack type generalization based semi-supervised protocol, WMCA~\cite{george2019biometric} with rich (7) attack types is utilized for evaluation. Specifically, different from the original leave-one-type-out testing on WMCA, here we adopt live and spoof samples with five attack types as the labeled training data while samples with another attack type is set as unlabeled data. Subsequently, the samples with the remaining one attack type are set as the target test set. Thus, to efficiently use such unlabeled data with unknown attack types is the key in this protocol.

\section{Experiments}

\begin{table}[t]
\centering
\caption{The ablation results on \emph{SSL-Protocol-1} on intra 20\% L+S of OULU-NPU. \emph{Cos.-Sim.} here denotes the cosine similarity loss between two global average pooled vectors.}
\vspace{-0.8em}
\resizebox{0.46\textwidth}{!}{
\begin{tabular}{l|c c c |c}
\toprule[1pt]
Module & Cos.-Sim. &Pred.-Consis.  &Embedd.-Consis.  &ACER(\%)\\
\hline
Model 1 &$\surd$  &\        &\         &4.38 \\
Model 2 &\        &$\surd$  &\         &3.54 \\ 
Model 3 &\        &\        &$\surd$   &2.29 \\
Model 4 &\        &$\surd$  &$\surd$   &\textbf{2.08} \\ 
\bottomrule[1pt]
\end{tabular}
}

\label{tab:ablation}
\end{table}

\begin{table}[t]
\centering
\caption{The ablation results on \emph{SSL-Protocol-1} on intra 20\% L+S of OULU-NPU. The two parts in Proof.~\ref{eq:dense_sim_v2_proof} are compared here.}
\vspace{-0.8em}
\resizebox{0.46\textwidth}{!}{
\begin{tabular}{l|c c|c}
\toprule[1pt]
Module &$\sqrt{\mathcal{DS}(\mathbf{H}, \mathbf{H}) \mathcal{DS}(\mathbf{F}, \mathbf{F})}$ &$cos<\Bar{\mathbf{H}},\Bar{\mathbf{F}}>$    &ACER(\%)\\
\hline
Model 3.1 &$\surd$  &\        &3.75 \\
Model 3.2 &\        &$\surd$  &3.75 \\ 
Model 3   &$\surd$  &$\surd$  &\textbf{2.29} \\
\bottomrule[1pt]
\end{tabular}
}

\label{tab:ablation_eqproof}
\end{table}

\begin{table}[t]
\centering
\caption{The comparison (ACER(\%)) among unsupervised methods w/ and wo/ EPCR (Embedd./Pred.-Consis.) on \emph{SSL-Protocol-1} on intra 20\% L+S of OULU-NPU. DenseCL~\cite{wang2020dense} is reproduced based on MoCov2~\cite{chen2020improved}, in which the dense contrastive learning module in DenseCL~\cite{wang2020dense} is plugged.}
\vspace{-0.8em}
\resizebox{0.3\textwidth}{!}{
\begin{tabular}{l|c c}
\toprule[1pt]
Module & original & w/ EPCR \\
\hline
MoCov2~\cite{chen2020improved}   &5.42    &2.92  \\
DenseCL~\cite{wang2020dense}   &4.38    &2.92  \\
BYOL~\cite{grill2020bootstrap}     &4.79    &2.50  \\
SimCLR~\cite{chen2020simple}   &5.00    &2.08  \\
SimSiam~\cite{chen2020exploring}  &4.38    &2.08  \\
\bottomrule[1pt]
\end{tabular}
}

\label{tab:ablation_unsupervised}
\vspace{-0.5em}
\end{table}

\subsection{Implementation Details}

\noindent \textbf{Training strategy.} 
The proposed method could be trained with an end-to-end strategy. Specifically, the labeled data forward and backward in the whole components of the network, which are supervised by the loss in Eq.~\ref{eq:loss_overall}. Meanwhile, the unlabeled data can not access labels, thus will not be supervised by $\mathcal{L}_{supervised}$ in Eq.~\ref{eq:loss_supervised}. Besides those augmentation methods in ~\cite{yu2020searching}, the patch shuffle augmentation~\cite{zhang2021structure} is also introduced as an effective augmentation method for our EPCR method.

\noindent \textbf{Testing strategy.}
The final spoof score is calculated by the mean of $\mathbf{C_{F_2}}$, which is generated by the bottom branch in Fig.~\ref{fig:overall_framework} consisting of a dense encoder $f$ and a classifier $c$.

\begin{table}[t]
\caption{The intra-dataset testing results on OULU-NPU. }
\vspace{-0.8em}
\resizebox{0.45\textwidth}{!}{
\begin{tabular}{c|c|c|c|c}
\toprule[1pt]
Prot. & Method & APCER(\%) & BPCER(\%) & ACER(\%) \\
\hline
\multirow{7}{*}{1} 
        &Disentangled ~\cite{zhang2020face} &1.7 &0.8 & 1.3 \\
        &FAS-SGTD~\cite{wang2020deep} &2.0 &0.0 & 1.0 \\
        &STDN~\cite{liu2020disentangling} &0.8 &1.3 & 1.1 \\
        &CDCN~\cite{yu2020searching} &0.4 &1.7 & 1.0 \\
        &BCN ~\cite{yu2020face} &0.0 &1.6 & \textbf{0.8} \\ \cline{2-5}
        &SimSiam-CDCN &0.4 &5.0 &2.7 \\ 
        &\textbf{EPCR(Ours)} &1.7 &0.0 & \textbf{0.8} \\
\midrule[1pt]
\multirow{7}{*}{2} 
       &STASN~\cite{yang2019face} &4.2 &0.3 & 2.2 \\
       &FAS-SGTD~\cite{wang2020deep} &2.5 &1.3 &1.9 \\
       &STDN~\cite{liu2020disentangling} &2.3 &1.6 & 1.9 \\
       &BCN ~\cite{yu2020face} &2.6 &0.8 & 1.7 \\
       &CDCN~\cite{yu2020searching} &1.5 &1.4 & 1.5 \\ \cline{2-5}
       &SimSiam-CDCN &3.1 &0.3 &1.7 \\
       &\textbf{EPCR(Ours)} &0.8 &0.3 & \textbf{0.6} \\
\midrule[1pt]
\multirow{7}{*}{3} 
       
       &STDN~\cite{liu2020disentangling} &1.6$\pm$1.6 &4.0$\pm$5.4 &{2.8}$\pm${3.3} \\
       &FAS-SGTD~\cite{wang2020deep} &3.2$\pm$2.0 &2.2$\pm$1.4 &{2.7}$\pm${0.6} \\
       &BCN ~\cite{yu2020face}&2.8$\pm$2.4 &2.3$\pm$2.8  &2.5$\pm$1.1 \\
       &CDCN~\cite{yu2020searching} &2.4$\pm$1.3 &2.2$\pm$2.0 &{2.3}$\pm${1.4} \\
       &Disentangled ~\cite{zhang2020face}&2.8$\pm$2.2 &1.7$\pm$2.6  &2.2$\pm$2.2 \\\cline{2-5}
       &SimSiam-CDCN &2.5$\pm$1.6 &3.1$\pm$2.5 &{2.8}$\pm${1.5} \\
       &\textbf{EPCR(Ours)} &0.4$\pm$0.5  &2.5$\pm$3.8  &\textbf{1.5}$\pm$\textbf{2.0} \\
\midrule[1pt]
\multirow{7}{*}{4} 
       &FaceDs~\cite{jourabloo2018face} &1.2$\pm$6.3 &6.1$\pm$5.1 &5.6$\pm$5.7 \\
       &BCN ~\cite{yu2020face}&2.9$\pm$4.0 &7.5$\pm$6.9  &5.2$\pm$3.7 \\
       &FAS-SGTD~\cite{wang2020deep} &6.7$\pm$7.5 &3.3$\pm$4.1 &5.0$\pm$2.2 \\
       &Disentangled ~\cite{zhang2020face}&5.4$\pm$2.9 &3.3$\pm$6.0  &4.4$\pm$3.0 \\
       &STDN~\cite{liu2020disentangling} &2.3$\pm$3.6 &5.2$\pm$5.4 &{3.8}$\pm${4.2} \\ \cline{2-5}
       &SimSiam-CDCN &0.8$\pm$2.0 &7.5$\pm$11.7 &{4.2}$\pm${6.8} \\
       &\textbf{EPCR(Ours)} &0.8$\pm$2.0 &5.8$\pm$8.0 &\textbf{3.3$\pm$4.9} \\
\bottomrule[1pt]
\end{tabular}
}

\label{tab:OULU}
\vspace{-0.3em}
\end{table}

\begin{table*}[!htb]
	\centering	
	\small
	\caption{Results on the DG cross-dataset testing protocols. The methods in the upper part are trained without domain information while those in the lower part are leveraging domain clues for generalization.}
	\vspace{-1.0em}
	\scalebox{0.9}{\begin{tabular}{c|c|c|c|c|c|c|c|c}
	\midrule[1pt]
		\multirow{2}{*}{\textbf{Method}} & \multicolumn{2}{c|}{\textbf{O\&C\&I to M}} & \multicolumn{2}{c|}{\textbf{O\&M\&I to C}} 
		& \multicolumn{2}{c|}{\textbf{O\&C\&M to I}}& \multicolumn{2}{c}{\textbf{I\&C\&M to O}}\\ \cline{2-9} 		
		& HTER(\%)                                   & AUC(\%)                                   & HTER(\%)                                   & AUC(\%)                                   & HTER(\%)                                   & AUC(\%)                                   & HTER(\%)                                   & AUC(\%)                                   \\ \hline
		Binary CNN~\cite{Yang2014Learn}				  &29.25&82.87  &34.88&71.94 	&34.47&65.88 	&29.61&77.54\\
		IDA~\cite{wen2015face}						  &66.67&27.86	&55.17&39.05	&28.35&78.25	&54.20&44.59\\
		Color Texture~\cite{Boulkenafet2017Face}      &28.09&78.47  &30.58&76.89    &40.40&62.78    &63.59&32.71\\
		LBP-TOP~\cite{de2014face}    				  &36.90&70.80  &42.60&61.05    &49.45&49.54    &53.15&44.09\\		
		Auxiliary(Depth)~\cite{Liu2018Learning}  &22.72&85.88	&33.52&73.15	&29.14&71.69	&30.17&77.61\\
		Auxiliary(All)~\cite{Liu2018Learning}		  &--&--		&28.4&--	    &27.6&--	    &--&--\\
		
		\textbf{EPCR(Ours)}               				  &\textbf{12.5}&\textbf{95.3} 	&\textbf{18.9}&\textbf{89.7}	&\textbf{14.0}&\textbf{92.4}		&\textbf{17.9}&\textbf{90.9} \\

		\hline
		MMD-AAE~\cite{li2018domain}					  &27.08&83.19 	&44.59&58.29	&31.58&75.18	&40.98&63.08\\			
		MADDG~\cite{shao2019multi}   				  &17.69&88.06 	&24.5&84.51	    &22.19&84.99 	&27.98&80.02 \\
		MDRL~\cite{wang2020cross}                     &17.02&90.01 	&19.68&87.43	&20.87&86.72	&25.02&81.47 \\
		SSDG-M~\cite{jia2020single}                   &16.67&90.47 	&23.11&85.45	&18.21&94.61	&25.17&81.83 \\
		RFM~\cite{shao2020regularized}                &13.89&93.98 	&20.27&88.16	&17.3&90.48	    &16.45&91.16 \\
		SDA~\cite{wang2021self}                       &15.4&91.8 	&24.5&84.4	    &15.6&90.1	    &23.1&84.3 \\
		
		\textbf{EPCR+SSDG(Ours)}               				  &\textbf{10.0}&\textbf{95.5} 	&\textbf{17.8}&\textbf{89.8}	&\textbf{13.5}&\textbf{96.0}	&\textbf{14.4}&\textbf{93.3} \\
		\bottomrule[1pt]
	\end{tabular}}
	
    \label{tab:DG}
    \vspace{-1.0em}
\end{table*}

\vspace{0.2em}
\noindent \textbf{Performance metrics.}
OULU-NPU uses 1) Attack Presentation
Classification Error Rate $APCER$, evaluating the highest error among all PAIs (e.g.,
print or display), 2) Bona Fide Presentation Classification Error Rate $BPCER$, evaluating the error of real access data, and 3) $ACER$~\cite{ACER}, evaluating the mean of $APCER$ and $BPCER$.
Following~\cite{shao2019multi}, the Half Total Error (HTER) and the Area Under Curve (AUC) are adopted as the evaluation metrics of DG.

\subsection{Experimental Results}
\begin{table}[t]
\centering
\caption{
The results of \emph{SSL-Protocol-1} on OULU-NPU.
For example, `intra 10\% L+S' denotes that 10\% live (L) and spoof (S) faces from original Protocol-1 are set labeled. And the remaining 90\% faces from original Protocol-1 are set unlabeled. We split the training set by the subject ID. \textbf{Supervised Only} means that EPCR is only supervised by the intra labeled data.}
\vspace{-0.8em}
\resizebox{0.48\textwidth}{!}{
\begin{tabular}{l|c c c c}
\toprule[1pt]
Protocols &Supervised Only &Mean-Teacher &USDAN &EPCR  \\
\hline
intra 10\% L+S  &10.42  &9.58  &15.63  &2.71 \\
intra 20\% L+S  &3.33   &5.62  &8.54   &2.08 \\
intra 50\% L+S  &2.71   &3.75  &7.50   &2.08\\
intra 80\% L+S  &1.67   &2.08  &6.45   &1.46\\
intra 100\% L+S &0.83   &1.88  &6.04     &0.83\\

\bottomrule[1pt]
\end{tabular}
}

\label{tab:semi_OULU_intra}
\vspace{-0.1em}
\end{table}

\subsubsection{Ablation Study}

Four models are implemented to demonstrate the effectiveness of important components in the proposed method. As shown in Tab.~\ref{tab:ablation}, Model 1 represents the original SimSiam (SimSiam-CDCN) whose backbone is CDCN and is supervised by the cosine similarity loss between the global average pooled embedding of two views. 
Model 2 is supervised with embedding-level consistency regularization with the proposed dense similarity loss. Model 3 is supervised with prediction-level consistency regularization. Model 4 denotes the model that trained with both \textbf{Embedd.-Consis.} and \textbf{Pred.-Consis.}. 

\textbf{Efficacy of consistency regularization.} 
As demonstrated in Tab.~\ref{tab:ablation}, with the progressive lower ACER of Model 2 and Model 3, it is clear that Embedd.-Consis. and Pred.-Consis. can provide valuable consistency regularization to learn discriminative features, respectively. Model 3 is superior to Model 2 with a large margin, which demonstrates that Embedd.-Consis. is most significant in our framework. Finally, compared with the other architectures, Model 4 obtains the lowest ACER, indicating that the combination of Embedd.-Consis. and Pred.-Consis. can further improve the performance of FAS. Furthermore, Tab.~\ref{tab:ablation_eqproof} shows that the two parts in Proof.~\ref{eq:dense_sim_v2_proof} are both important in the Embedd.-Consis. and can be complementary to each other.

\textbf{Comparison with vanilla SimSiam.}
As shown in Tab.~\ref{tab:OULU}, we also conduct vanilla SimSiam~\cite{chen2020exploring} with the CDCN backbone. It can be seen that the ACER of SimSiam-CDCN is much higher than that of the proposed EPCR, and even higher than the ACER of the original CDCN in three of four protocols of OULU-NPU. This demonstrates that a pure SimSiam cannot perform well on FAS tasks. We recognize the reason is that the global averaged features and vector similarity damage the learning of detailed spoof clues. 

\textbf{EPCR on recent contrastive learning based unsupervised structures.}
Tab.~\ref{tab:ablation_unsupervised} shows that the performance of recent unsupervised structures with Embedd.-Consis. and Pred.-Consis. can outperform that of those original methods, indicating that 
the proposed combination of Embedd.-Consis. and Pred.-Consis. are the core factors in our EPCR framework, which can be plugged into any contrastive learning based unsupervised structure for improvement. In particular, the ACER of DenseCL~\cite{wang2020dense} is superior to that of Mocov2~\cite{chen2020improved}, but is inferior to that of the EPCR. This implies that DenseCL~\cite{wang2020dense} can indeed improve the performance of FAS models. However, our proposed dense similarity loss can provide more suitable consistency regularization for face anti-spoofing instead of the dense contrastive learning module in DenseCL~\cite{wang2020dense}.

\vspace{-0.8em}
\subsubsection{Experiments on Supervised FAS}
We conduct experiments on fully-supervised FAS tasks. Original four protocols in OULU-NPU and domain generation (DG) cross-dataset testing protocols are used here. 

\textbf{Results on OULU-NPU}. \quad Tab.~\ref{tab:OULU} shows that the proposed method ranks first on all 4 protocols. In protocols 2, 3, and 4, compared with the state-of-the-art methods, the EPCR can relatively reduce the ACER(\%) by 60.0\%, 31.8\%, and 13.2\%, respectively. Especially in protocol 2 and 3, our method is the first to reach ACER(\%) lower than 1.0 and 2.0, respectively. The results on OULU-NPU indicate the proposed method performs well at the generalization of the external environment, attack mediums, and input camera variations. 

\textbf{Results on DG}. Tab.~\ref{tab:DG} compares the performance of our method with the state-of-the-art methods on DG protocols. We can see that our proposed method ranks first among all models trained without domain information. EPCR can also surpass the previous models trained by leveraging domain clues in [O\&C\&I to M] and [O\&M\&I to C] and ranks second in [O\&C\&M to I] and [I\&C\&M to O]. The comparable results in standard DG evaluation demonstrate the superiority of the proposed EPCR in domain generalization. When joined with SSDG~\cite{jia2020single}, EPCR can improve the performance slightly, which implies that the proposed EPCR is complementary to the method using domain information.

\vspace{-0.8em}
\subsubsection{Experiments on Semi-supervised FAS}
\label{sec:semiExeperiment}


The semi-supervised FAS protocols have been demonstrated in the previous section.

\vspace{0.3em}
\emph{SSL-Protocol-1}: It can be seen from Tab.~\ref{tab:semi_OULU_intra} that with the increase of labeled samples, the performance of FAS methods is getting better. EPCR with unlabeled data outperforms that without unlabeled data (supervised only) in each level, which indicates that unlabeled data is beneficial for the model's generalization. And our proposed EPCR can reach lower ACER(\%) than Mean-Teacher and USDAN in all sub-protocols with different labeled/unlabeled proportions, indicating the effectiveness of the proposed EPCR.

\vspace{0.3em}
\emph{SSL-Protocol-2}: Tab.~\ref{tab:semi_OULU_extra} shows that the presented SSL methods can make great progress with extra unlabeled data which contains similar domain information with target data. It is worth noting that merely introducing extra live faces could still improve the performance. The experimental phenomenon can bring insight for real-world FAS applications that collecting enough live data with the target domain can effectively promote the generalization capacity on the target domain.

\emph{SSL-Protocol-3}: As shown in Tab.~\ref{tab:semi_DG_leaveoneout}, we find that EPCR with the extra unlabeled dataset can outperform other baselines in four of six sub-protocols (i.e., [O\&I to C], [O\&M to I], [C\&M to O] and [C\&I to O]), and even gets better performance than the model with extra labeled, which demonstrates that open-world data with labels may bring harmful impacts to the model's generalization. The possible reason is that the labeled data with different domains from the target domain will add noise in the model training.

\begin{table}[t]
\centering
\caption{The results of \emph{SSL-Protocol-2} based on OULU-NPU. For example, besides all labeled data from the original Protocol 1, we also take 50\% live (L) and spoof (S) faces from the whole training set as unlabeled data. This situation is named `extra 50\% L+S'. }
\vspace{-0.8em}
\resizebox{0.48\textwidth}{!}{
\begin{tabular}{l|c c c c}
\toprule[1pt]
Protocols &Supervised Only &Mean-Teacher  &USDAN  &EPCR  \\
\hline
extra 0\%        &0.83    &1.88   &6.04   &0.83  \\
extra 50\% L     &0.83    &2.08   &5.63   &0.42  \\
extra 100\% L    &0.83    &0.63   &5.63   &0.21  \\
extra 50\% L+S   &0.83    &0.42   &4.37   &0.21  \\
extra 100\% L+S  &0.83    &0.00   &2.29   &0.00  \\

\bottomrule[1pt]
\end{tabular}
}

\label{tab:semi_OULU_extra}
\vspace{-0.4em}
\end{table}

\begin{table*}[!htb]
	\centering	
	\small
	\caption{Results of \emph{SSL-Protocol-3} with \textbf{leave-one-unlabeled semi-supervised} setting. For example, in the sub-protocl [O\&I to M], data from the remaining dataset `C' (CASIA-MFSD) are used as extra labeled or unlabeled training data. }
	\vspace{-0.8em}
	\scalebox{0.9}{\begin{tabular}{l|c|c|c|c|c|c|c|c|c|c|c|c}
		\toprule[1pt]
		\multirow{2}{*}{\textbf{Method}} & \multicolumn{2}{c|}{\textbf{O\&I to M}} & \multicolumn{2}{c|}{\textbf{M\&I to C}} 
		& \multicolumn{2}{c|}{\textbf{O\&I to C}}& \multicolumn{2}{c|}{\textbf{O\&M to I}}& \multicolumn{2}{c|}{\textbf{C\&M to O}}& \multicolumn{2}{c}{\textbf{C\&I to O}}  \\ \cline{2-13} 		
		& HTER    & AUC   & HTER    & AUC    & HTER    & AUC   & HTER    & AUC   & HTER    & AUC   & HTER    & AUC                                \\ \hline
    
    Supervised Only &12.1&94.2    &30.4&77.0   &18.0&90.1   &16.8&93.8   &17.9&89.5   &15.9&91.9 \\
    
    Mean-Teacher~\cite{tarvainen2017mean} w/ extra unlabeled &19.6&86.5    &31.1&76.6   &23.7&84.9   &18.4&86.0   &23.5&84.9   &19.9&88.2 \\
    
    USDAN~\cite{jia2021unified} w/ extra unlabeled  &15.8&88.1    &35.6&69.0   &33.3&72.7   &19.8&87.9   &20.2&88.3   &27.5&81.9 \\
    
    EPCR  w/ extra labeled  &12.5&\textbf{95.3}    &\textbf{18.9}&\textbf{89.7}    &18.9&89.7    &14.0&92.4   &17.9&90.9   &17.9&90.9 \\
    
    EPCR  w/ extra unlabeled  &\textbf{10.4}&94.5    &25.4&83.8   &\textbf{16.7}&\textbf{91.4}   &\textbf{12.4}&\textbf{94.3}   &\textbf{17.8}&\textbf{91.3}   &\textbf{14.7}&\textbf{92.1} 
    
    \\
	\bottomrule[1pt]	
	\end{tabular}}
	
    \label{tab:semi_DG_leaveoneout}
    \vspace{-0.2em}
\end{table*}

\begin{table*}[!htb]
	\centering	
	\small
	
\caption{Results of \emph{SSL-Protocol-4} with \textbf{partly intra-labeled semi-supervised} setting. We set respective 20\%, 50\% and 100\% subjects from the standard DG protocols as labeled data while the remaining data from other subjects are set unlabeled. }
\vspace{-0.8em}	
	\scalebox{0.95}{\begin{tabular}{l|c|c|c|c|c|c|c|c|c|c|c|c}
		\toprule[1pt]
		\multirow{2}{*}{\textbf{Method} \quad with AUC(\%)} & \multicolumn{3}{c|}{\textbf{O\&C\&I to M}} & \multicolumn{3}{c|}{\textbf{O\&M\&I to C}} 
		& \multicolumn{3}{c|}{\textbf{O\&C\&M to I}}& \multicolumn{3}{c}{\textbf{I\&C\&M to O}} 
		\\ \cline{2-13} 		
		& 20\%   & 50\%  & 100\%  & 20\%    & 50\%   & 100\%   & 20\%    & 50\%  & 100\%  & 20\%    & 50\%  & 100\%                           
		\\ \hline
    
    Supervised Only             &86.9  &93.2  &95.3  &88.5  &88.5  &89.7  &83.5  &84.9  &92.4  &82.4  &88.9  &90.9\\
    Mean-Teacher~\cite{tarvainen2017mean} w/ unlabeled   &89.1  &93.9  &--    &86.9  &86.8  &--    &84.0  &87.0  &--    &85.0  &90.5  &--\\
    USDAN~\cite{jia2021unified} w/ unlabeled   &82.3  &89.6  &--    &75.1  &75.4  &--    &87.8  &89.7  &--    &85.2  &92.6  &--\\
    EPCR w/ unlabeled       &90.0  &95.6  &--    &91.0  &90.8  &--    &87.3  &89.8  &--    &85.2  &92.7  &--         
    
    \\
	\bottomrule[1pt]	
	\end{tabular}}
	
    \label{tab:semi_DG_intra}
    \vspace{-0.2em}
\end{table*}

\begin{table*}[!htb]
\centering
\caption{Results of \emph{SSL-Protocol-5} with one `Unseen' attack type (with \emph{Italics}) for unlabeled training while the other one (without \emph{Italics}) for testing on WMCA~\cite{george2019biometric}. 
`-SL' and `-SSL' represents fully-supervised and semi-supervised training, respectively. 
}
\vspace{-0.8em}
\scalebox{0.77}{
\begin{tabular}{c|c|cccccccc}
\toprule[1pt]
\multirow{2}{*}{Modality} & \multirow{2}{*}{Method} & \multicolumn{8}{c}{Unseen attack type}                                                                                                                  \\ \cline{3-10} 
                          &                         & Flexiblemask  & Replay        & Fakehead       & Prints         & Glasses        & Papermask     & Rigidmask      & Mean$\pm$Std            \\ \hline
& MC-PixBiS~\cite{george2019deep}                                 & 49.7          & 3.7           & 0.7            & 0.1            & 16.0           & 0.2           & 3.4            & \textbf{10.5$\pm$16.7}           \\
        RGB-Depth                & MCCNN-OCCL-GMM~\cite{george2020learning}                            & 22.8          & 31.4          & 1.9            & 30.0           & 50.0           & 4.8           & 18.3           & 22.74$\pm$15.3          \\
           -SL                & MC-ResNetDLAS~\cite{parkin2019recognizing}                             & 33.3          & 38.5          & 49.6           & 3.80           & 41.0           & 47.0          & 20.6           & 33.4$\pm$14.9           \\ \hline
\multirow{3}{*}{RGB-SL}      
                          & Resnet50~\cite{he2016deep}         & 16.13          & 32.88         & 14.36   & 0.03          & 45.71          & 12.09         & 12.19    & 20.09$\pm$14.91   
                          \\
                          & CDCN~\cite{yu2020searching}         & 10.39          & 22.67         & 3.70   & 0.49          & 43.08          & 3.25         & 2.76 & 12.33$\pm$15.52   
                          \\
                          & EPCR         & 9.91          & 20.89         & 4.97 & 0.28         & 41.61          & 2.47         & 2.78    & \textbf{11.85$\pm$14.85}   
                          \\ \hline
\multirow{5}{*}{RGB-SSL} 
        &  &\emph{Papermask}      &\emph{Prints}          &\emph{Rigidmask}        &\emph{Papermask}           &\emph{Papermask}        &\emph{FlexibleMask}      &\emph{Fakehead}  & 
        \\ \cline{3-10}
        &Supervised Only       &12.26          &25.74           &6.30           &0.12      &44.93             &1.00            &2.22     &13.23$\pm$16.60
        \\
        &Mean-Teacher~\cite{tarvainen2017mean}       &10.83          &30.55          &5.95           &0.09      &42.45             &4.42            &5.15     &14.20$\pm$15.92
        \\
        &USDAN~\cite{jia2021unified}       &13.46          &26.03          &7.15           &1.83      &34.95             &4.79            &10.51     &14.10$\pm$12.08
        \\
        &EPCR       &12.42          &22.16           &4.21           &0.05      &43.21             &2.08            &4.03     &\textbf{12.61$\pm$15.53}
        \\
       \bottomrule[1pt]

\end{tabular}
}

\vspace{-0.3em}
\label{tab:semi_WMCA}
\end{table*}

\emph{SSL-Protocol-4}: As illustrated in Tab.~\ref{tab:semi_DG_intra}, semi-supervised EPCR and Mean-Teacher both obtain higher AUC than the EPCR without unlabeled data (supervised only), and our proposed semi-supervised EPCR can achieve the best performance in seven of eight SSL sub-protocols. Moreover, we find that there is very little or no improvement from 20\% labeled data to 50\% labeled data in the sub-protocol [O\&M\&I to C], which might be explained that the extra labeled data harms the domain generalization. According to  Tab.~\ref{tab:semi_DG_leaveoneout} and Tab.~\ref{tab:semi_DG_intra}, we discover that the combination between semi-supervised learning and domain generalization is a significant direction for future FAS research.

\begin{figure}[!t]
\vspace{-0.2em}
\centering
  \includegraphics[width=0.45\textwidth]{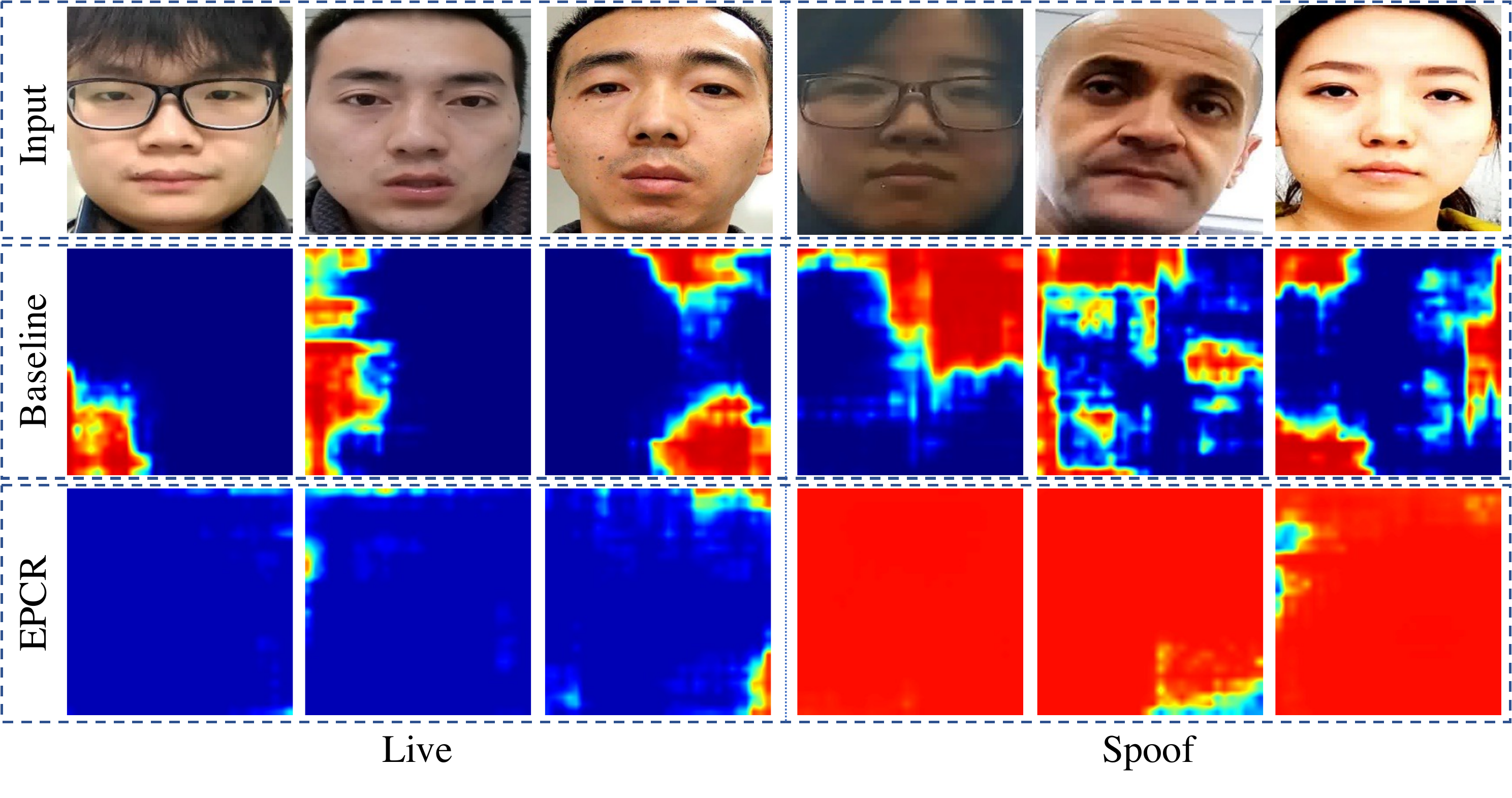}
  \vspace{-1.0em}
  \caption{
    Visualization of hard samples on Protocol-2 of OULU-NPU. The predicted maps from CDCN baseline and EPCR are shown in the second and third row, respectively.
  }
  \label{fig:res_maps}
  \vspace{-1em}
\end{figure}

\emph{SSL-Protocol-5}: The samples whose attack type is most similar to that of testing data are selected as the unlabeled training data. As shown in Tab.~\ref{tab:semi_WMCA}, in `RGB-SSL' part, EPCR with extra unlabeled data (EPCR) outperforms EPCR without extra unlabeled data (Supervised Only), which implies that introducing extra unlabeled and unseen attack data can promote the effectiveness of models. Moreover, the proposed EPCR can obtain state-of-the-art performance in both `RGB-SL' and `RGB-SSL' parts. 

\subsubsection{Visualization and Analysis}
The predicted maps for several hard samples are visualized as heatmaps in Fig.~\ref{fig:res_maps}. It can be seen that the spoof samples in the last three columns are difficult for the raw CDCN baseline to distinguish. Meanwhile, the generated maps of the baseline for the live samples present noise in several regions. In contrast, the proposed EPCR provides preciser predictions than the baseline, indicating the discriminative feature representation capacity of the proposed EPCR.

\section{Conclusion}
In this paper, we investigate that consistency regularization is significant for the reliability of models. Inspired by this motivation, we propose a novel embedding-level and prediction-level consistency regularization method for deep face anti-spoofing, which performs well on both full-supervised and semi-supervised FAS tasks. And a dense similarity objective is designed to exploit more effective consistency regularization of FAS. We also provide five novel semi-supervised learning benchmarks with diversity for the FAS research community. In the future, we will exploit consistency regularization formulation on other dimensions, such as temporal consistency along with different frames in one facial video.


{\small
\bibliographystyle{ieee_fullname}
\bibliography{egbib}
}

\clearpage

\section*{A. More Analysis of Dense Similarity}
We first review the similarity in previous contrastive learning~\cite{chen2020exploring}. Given two 
flattened tensors $\mathbf{H}, \mathbf{F} \in \mathcal{R}^{s^2\times d}$, the cosine similarity is calculated between the global averaged  $\mathbf{H}$ and $\mathbf{F}$ as below:
\begin{equation}
\mathcal{S}(\mathbf{H}, \mathbf{F})  =  cos<\Bar{\mathbf{H}}',\Bar{\mathbf{F}}'>,
\label{eq:traditional_sim}
\end{equation}
where $\Bar{\mathbf{H}}'$ (or $\Bar{\mathbf{F}}'$) is the global representation of all row vectors in $\mathbf{H}$ (or $\mathbf{F}$)

Considering the most common presentation attack instruments (i.e., print attack, replay attack, and 3D mask) in FAS, each local patch in the facial region might share the same spoof clues (e.g., material and quality).
Therefore, we can reasonably assume that the vector at each point of $\mathbf{H_1}$ contains the maximal similarity to that of $\mathbf{F_2}$, named as \textbf{\emph{dense similarity}}.
The dense similarity between $\mathbf{H}$ and $\mathbf{F}$ is defined as below:
\begin{equation}
\mathcal{DS}(\mathbf{H}, \mathbf{F}) = \sum_{i} \sum_{j} \lnorm{\mathbf{H}^{i}}{\cdot}\lnorm{\mathbf{F}^{j}},
\label{eq:dense_sim}
\end{equation}
where $\mathbf{H}^{i}$ denotes the $i$-th row of $\mathbf{H}$, and $\cdot$ denotes the dot product. 

\begin{lemma}
The property of $\mathcal{DS}$ can be derived as below:
\begin{equation}
\mathcal{DS}(\mathbf{H}, \mathbf{F})  = \sqrt{\mathcal{DS}(\mathbf{H}, \mathbf{H}) \mathcal{DS}(\mathbf{F}, \mathbf{F})} cos<\Bar{\mathbf{H}},\Bar{\mathbf{F}}>,
\label{eq:dense_sim_v2}
\end{equation}
where $\Bar{\mathbf{H}}$ (or $\Bar{\mathbf{F}}$) is the center of all normalized row vectors in $\mathbf{H}$ (or $\mathbf{F}$).
\end{lemma}

\begin{proof}
As $\Bar{\mathbf{H}}$ is the center of all normalized row vectors in $\mathbf{H}$, then,
\begin{scriptsize}
\begin{equation*}
\Bar{\mathbf{H}} = \frac{1}{s^2}\sum_{i}\lnorm{\mathbf{H}^{i}},
\end{equation*}
\end{scriptsize}

\begin{scriptsize}
\begin{equation*}
\begin{split}
& \mathcal{DS}(\mathbf{H}, \mathbf{F})  \\ 
& = \sum_{i} \sum_{j} \lnorm{\mathbf{H}^{i}}{\cdot}\lnorm{\mathbf{F}^{j}} \\
& = \sum_{i}\lnorm{\mathbf{H}^{i}} {\cdot} \sum_{j}\lnorm{\mathbf{F}^{j}} \\
& = \twonorm{\sum_{i}\lnorm{\mathbf{H}^{i}}} \twonorm{\sum_{j}\lnorm{\mathbf{F}^{j}}} cos<\sum_{i}\lnorm{\mathbf{H}^{i}}, \sum_{j}\lnorm{\mathbf{F}^{j}}> \\
& =  \twonorm{\sum_{i}\lnorm{\mathbf{H}^{i}}} \twonorm{\sum_{j}\lnorm{\mathbf{F}^{j}}} cos<\Bar{\mathbf{H}},\Bar{\mathbf{F}}> \\ 
& = \sqrt{\sum_{i}\lnorm{\mathbf{H}^{i}} \cdot \sum_{i}\lnorm{\mathbf{H}^{i}}}  \sqrt{\sum_{j}\lnorm{\mathbf{F}^{j}} \cdot \sum_{j}\lnorm{\mathbf{F}^{j}}} cos<\Bar{\mathbf{H}},\Bar{\mathbf{F}}> \\
& = \sqrt{\sum_{i}\lnorm{\mathbf{H}^{i}} \cdot \sum_{j}\lnorm{\mathbf{H}^{j}}}  \sqrt{\sum_{i}\lnorm{\mathbf{F}^{i}} \cdot \sum_{j}\lnorm{\mathbf{F}^{j}}} cos<\Bar{\mathbf{H}},\Bar{\mathbf{F}}> \\
& = \sqrt{\mathcal{DS}(\mathbf{H}, \mathbf{H})} \sqrt{\mathcal{DS}(\mathbf{F}, \mathbf{F})} cos<\Bar{\mathbf{H}},\Bar{\mathbf{F}}>.
\label{eq:dense_sim_v2_proof}
\end{split}
\end{equation*}
\end{scriptsize}

\end{proof}

Comparing Eq.~\ref{eq:dense_sim_v2} with Eq.~\ref{eq:traditional_sim}, we find that our dense similarity not only reflects the global relation between the two tensors by $ cos<\Bar{\mathbf{H}},\Bar{\mathbf{F}}>$ ( $cos<\Bar{\mathbf{H}}',\Bar{\mathbf{F}}'>$), but also considers the intra-tensor consistency in $\mathcal{DS}(\mathbf{H}, \mathbf{H})$ and $\mathcal{DS}(\mathbf{F}, \mathbf{F})$. This statement means that every point in $\mathbf{H}$ and $\mathbf{F}$ will be consistent with the global feature $\Bar{\mathbf{H}}$ (or $\Bar{\mathbf{F}}$), which explains the difference between dense similarity and previous similarity method in another form. 

\section*{B. Implementation Details}
\textbf{Baseline Settings.}
We adopt SGD optimizer with weight decay $0.0001$ and momentum $0.9$. The learning rate is computed by the formulation $lr\times$batch-size$/256$. And a cosine decay schedule~\cite{chen2020simple}
is implemented from the starting base $lr=0.03$ to the ending base $lr=0.01$. The batch size is 64 by default, which can be friendly conducted on eight $1080$Ti GPUs. Different from ~\cite{chen2020exploring}, we don't use synchronized batch normalization (BN) but standard BN in our experiment, and we find that synchronized BN brings a little damage to the performance. The possible reason is that large batch size when synchronizing BN will result in local optimal solution~\cite{masters2018revisiting}, especially for the FAS task in our experiments. 

\vspace{0.2em}

\textbf{Projector}. The projector $p$ contains three \emph{Conv-BN-ReLU} blocks, where each block consists of one $1\times1$ convolution layer with 64 dimensions, one BN layer and Relu layer. Specially, The last output block has no Relu. 

\vspace{0.2em}
\textbf{Predictor}. The predictor $h$ contains one \emph{Conv-BN-ReLU} block, which is cascaded with a single $1\times1$ convolution layer with 64 dimensions.

\textbf{Pseudo Code.} The pseudo code is shown in Algorithm.~\ref{alg:code}.

\begin{algorithm}[t]
\caption{Pseudo code for EPCR, PyTorch-like}
\label{alg:code}
\definecolor{codeblue}{rgb}{0.25,0.5,0.5}
\definecolor{codekw}{rgb}{0.85, 0.18, 0.50}
\lstset{
  backgroundcolor=\color{white},
  basicstyle=\fontsize{7.5pt}{7.5pt}\ttfamily\selectfont,
  columns=fullflexible,
  breaklines=true,
  captionpos=b,
  commentstyle=\fontsize{7.5pt}{7.5pt}\color{codeblue},
  keywordstyle=\fontsize{7.5pt}{7.5pt}\color{codekw},
}

\begin{lstlisting}[language=python]
# f: dense backbone + projector
# h: dense predictor
# c: dense classifier

# load a minibatch x with t samples and
# labels y (only for labeled data)
for x, y in loader:  
    x1, x2 = aug(x), aug(x)  # random augmentation
    # shape = (t, d, s, s)
    F1, F2 = f(x1), f(x2)  # features
    H1, H2 = h(F1), h(F2)  # embeddings
    # shape = (t, 1, s, s)
    CF1, CF2 = c(F1), C(F2) # predicted maps
    CH1, CH2 = c(H1), C(H2) # predicted maps
    
    # embedding-level consistency regularization
    L_embedd = D_dense(H1, F2)/2 
                + D_dense(H2, F1)/2   
    # prediction-level consistency regularization
    L_pred = MSE(CH1, CF2)/2 + MSE(CH2, CF1)/2
    # select labeled tensors
    CF1 = select_labeled_tensors(CF1) 
    CF2 = select_labeled_tensors(CF2) 
    # expand the label from scalars to maps
    Y = y.repeat(1, 1, s, s) 
    # standard supervised loss for labeled data
    L_supervised = MSE(CF1, Y)/2 + MSE(CF2, Y)/2
    # overall loss
    L =  L_supervised + L_embedd + \alpha * L_pred

    L.backward()  # back-propagate
    update(f, h, c)  # SGD update

def D_dense(H, F):  # dense similarity loss
    F = F.detach()  # stop gradient

    H = reshape(H, (t, s*s, d))
    F = reshape(F, (t, d, s*s))
    
    H = normalize(H, dim=2) # # l2-normalize
    F = normalize(F, dim=1) # # l2-normalize
    
    # batch matrix-matrix product of matrices 
    loss = bmm(H, F)

    return -loss.mean()
\end{lstlisting}
\end{algorithm}

\section*{C. Datasets and Protocols}
\subsection*{C-1. Datasets}
\textbf{OULU-NPU}~\cite{Boulkenafet2017OULU} consists of 4950 real access and spoofing videos with a high resolution. There are four official protocols to verify the generalization of face anti-spoofing (FAS) models. In particular, Protocol 1 and Protocol 2 are designed to validate the generalization of FAS methods under unseen  environmental condition, and  unseen attack medium (e.g., unseen printers or displays), respectively. In order to study the effect of the input camera variation, Protocol 3 leverages a Leave One Camera Out (LOCO) protocol. Protocol 4 integrates all the above factors from protocol 1 to 3, which is the most challenging.

\textbf{CASIA-MFSD}~\cite{Zhang2012A} contains 50 genuine subjects with total 600 videos. And three imaging qualities are considered, namely the low quality, normal quality, and high quality. There are three kinds of presentation attacks including warped photo attack, cut photo attack and video attack.

\textbf{Replay-Attack}~\cite{ReplayAttack} consists of short video recording of both real and fake faces, composed of 50 different identities. The video sequence in Replay-Attack~\cite{ReplayAttack} is  low-resolution, which is collected with a resolution of 320 by 240 pixels.

\textbf{MSU-MFSD}~\cite{wen2015face} is composed of 70 genuine videos and 210 spoof videos, which contains 35 subjects. The type of presentation attack in MSU-MFSD~\cite{wen2015face} is printed photo and replayed video. And the dataset involves a widely age range from 20 to 60.

\textbf{WMCA}~\cite{george2019biometric} contains data captured using multiple capturing devices/channels. The channels present are color, depth, thermal and infrared. And there are a various kinds of 2D and 3D presentation attacks, specifically, 3D print and replay attacks, mannequins, paper masks, silicon masks, rigid masks, transparent masks, and non-medical eyeglasses.

\subsection*{C-2. Protocols}
\textbf{\emph{SSL-protocol-1.}} There are three sessions based on different environmental conditions in Protocol 1 of OULU-NPU. For standard Protocol 1, the train and dev set consist of the data with the first two sessions (Session 1 and 2) while the test set only consists of the data with the third session (Session 3). In \emph{SSL-protocol-1}, we adopt specific ratio of subjects in Session 1 and 2 as the labeled train set, and the remaining subjects in Session 1 and 2 are regarded as unlabeled train set. For example, ``20\% intra live (L) and spoof (S)'' means that the top 20\% of the data in ascending order of subject ID is labeled and the last 80\% of the data in ascending order of subject ID is unlabeled. Note that the dev set and test set remain unchanged.

\textbf{\emph{SSL-protocol-2.}} Contrary to \emph{SSL-protocol-1}, \emph{SSL-protocol-2} utilizes all the data of Session 1 and 2 in the train set. Parts of subjects in Session 3 are also taken as unlabeled train data to introduce the same domain to the target domain of test set. For example, ``50\% extra live (L)'' means that the top 50\% of the live data in ascending order of subject ID is set as unlabeled data to be trained in the semi-supervised FAS task. In this protocol, we deliberately consider to validate the performance of models when only introducing extra live faces, which is more relevant to the practical situation.

\textbf{\emph{SSL-protocol-3.}} Considering that the data in OULU-
NPU is collected by researchers with particular behaviors, we try to conduct more challenging semi-supervised tasks on the domain generalization (DG)~\cite{shao2019multi} based FAS. In this protocol, we conduct experiments based on \textbf{leave-one-unlabeled} manner, which means that 
one of the three training sets from standard DG protocols as the unlabeled set and the other two as the labeled sets. For example, in the sub-protocol [O\&I to M], data from the remaining dataset ‘C’ (CASIA-MFSD) are used as extra labeled or unlabeled training data. In other words, this protocol can validate the performance of FAS models when the introduced data contain different domain from both the labeled training data and the testing data.

\textbf{\emph{SSL-protocol-4.}} Similar to standard DG evaluation, \emph{SSL-protocol-4} utilizes all three training set, but merely selects part of data in each training set as labeled data, and the remaining ones are regarded as unlabeled data. For example, ``20\%'' in [O\&C\&I to M] means that 20\% of the data in all three training set (O\&C\&I) is labeled  and the remaining 80\% of the data is unlabeled. Note that the labeled and unlabeled data are split by the subject ID, which is similar to \emph{SSL-protocol-1} and \emph{SSL-protocol-2}.

\textbf{\emph{SSL-protocol-5.}} In this attack type generalization based semi-supervised protocol, WMCA~\cite{george2019biometric} with rich (7) attack types is utilized for evaluation. Specifically, different from the original leave-one-type-out testing on WMCA, here we adopt live and spoof samples with five attack types as the labeled training data while samples with another attack type is set as unlabeled data. Subsequently, the samples with the remaining one attack type are set as the target test set. Thus, to efficiently use such unlabeled data with unknown attack types is the key in this protocol. For example, samples with `Live' and 5 attack types (`Replay', `Fakehead', `Prints', `Glasses' and `Rigidmask') are set as labeled training set, while samples with another one attack type (`Papermask') are set as unlabeled training set. Finally, the samples with the remaining one attack type (`Flexiblemask') are set as the target test set. In this protocol, the spoof pattern in `Flexiblemask' is similar to that in `Papermask'.

\section*{D. Comparison with More Baselines}

\begin{table}[t]
\centering
\caption{The study on more baselines on 20\% intra live (L) and spoof (S) of \emph{SSL-protocol-1}. ACER(\%) is used as the metric.}
\resizebox{0.49\textwidth}{!}{
\begin{tabular}{c c c c c}
\toprule[1pt]
Fixmatch~\cite{sohn2020fixmatch}  &Mixmatch~\cite{berthelot2019mixmatch}  &Mean-Teacher~\cite{tarvainen2017mean} &Supervised Only &EPCR \\
\hline
 11.04   &6.25   &5.62  &3.33   & 2.08 \\
\bottomrule[1pt]
\end{tabular}
}
\label{tab:ablation_more_baselines}
\end{table}







As shown in Tab.~\ref{tab:ablation_more_baselines}, besides Mean-Teacher~\cite{tarvainen2017mean}, we also carry out the experiments of more recent semi-supervised methods, i.e., Fixmatch~\cite{sohn2020fixmatch} and Mixmatch~\cite{berthelot2019mixmatch} on ``intra 20\% L+S" of the \emph{SSL-protocol-1}. It can be seen from Tab.~\ref{tab:ablation_more_baselines} that EPCR with extra labeled data  outperforms other baselines. Specially for EPCR supervised only with the labeled data, the ACER is lower than those of the previous three SSL methods, which demonstrates the effectiveness of the proposed consistency regularization method. The ACER of Mixmatch~\cite{berthelot2019mixmatch} is higher than that of Mean-Teacher~\cite{tarvainen2017mean}, which can be explained that mix-up operation will disrupt the spoof cues of facial images. Furthermore, we find that Fixmatch~\cite{sohn2020fixmatch} is inferior to Mean-Teacher~\cite{tarvainen2017mean} in this experiment, which indicates that the guessed label (the mask for the consistency regularization) by the prediction may be not suitable for the current FAS methods. Instead, a simple consistency regularization between the outputs of two augmented views~\cite{tarvainen2017mean} is an effective choice. Based on the above experimental phenomenon, we select Mean-Teacher~\cite{tarvainen2017mean} as the baseline SSL method in the most of SSL experiments for simplicity.

\section*{E. Data Augmentation}
\begin{figure}[!htb]
\centering
  \includegraphics[width=0.46\textwidth]{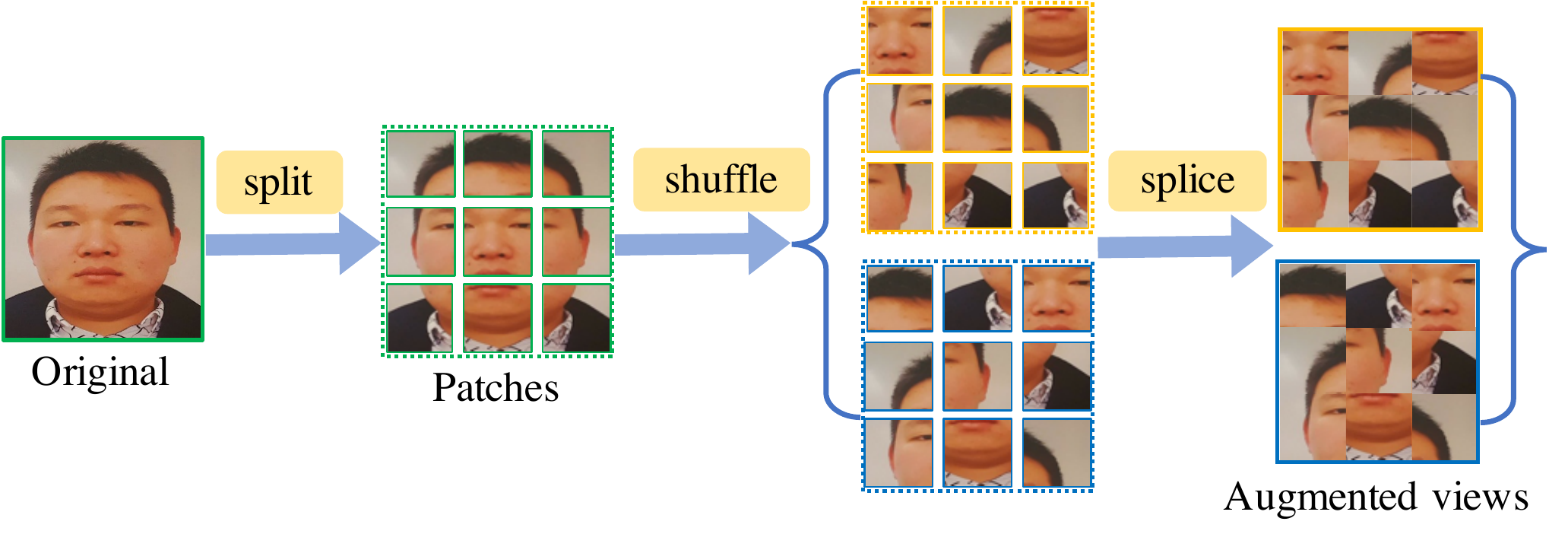}
   \vspace{-0.8em}
  \caption{
    PSA. The input is evenly split into nine patches, which are then shuffled and spliced to form the augmented views.
  }
  \label{fig:spa}
   \vspace{-0.8em}
\end{figure}

\begin{table}[t]
\centering
\caption{The study on different augmentations. The experiments are conducted on ``intra 20\% L+S" of the \emph{SSL-protocol-1}.}
\resizebox{0.49\textwidth}{!}{
\begin{tabular}{c c c c c c c c |c}
\toprule[1pt]
Crop     &Color    &Flip     &Cutout   &Blur     &PSA $2 \times 2$  &PSA $3 \times 3$   &PSA $4 \times 4$ &ACER(\%)\\
\hline
$\surd$  &         &         &         &         &         &         &            &7.71  \\
$\surd$  &$\surd$  &         &         &         &         &         &            &6.25  \\
$\surd$  &$\surd$  &$\surd$  &         &         &         &         &            &4.58  \\
$\surd$  &$\surd$  &$\surd$  &$\surd$  &         &         &         &            &3.54  \\
$\surd$  &$\surd$  &$\surd$  &$\surd$  &$\surd$  &         &         &            &4.79  \\
\midrule[1pt]
$\surd$  &$\surd$  &$\surd$  &$\surd$  &         &$\surd$  &         &            &4.17  \\
$\surd$  &$\surd$  &$\surd$  &$\surd$  &         &         &$\surd$  &            &\textbf{2.08}  \\
$\surd$  &$\surd$  &$\surd$  &$\surd$  &         &         &         &$\surd$  &3.33  \\

\bottomrule[1pt]
\end{tabular}
}
\label{tab:ablation_augmentation}
\end{table}

\begin{figure}[ht]
  \centering
  \begin{subfigure}[b]{0.141\textwidth}
    \includegraphics[width=1.0\linewidth]{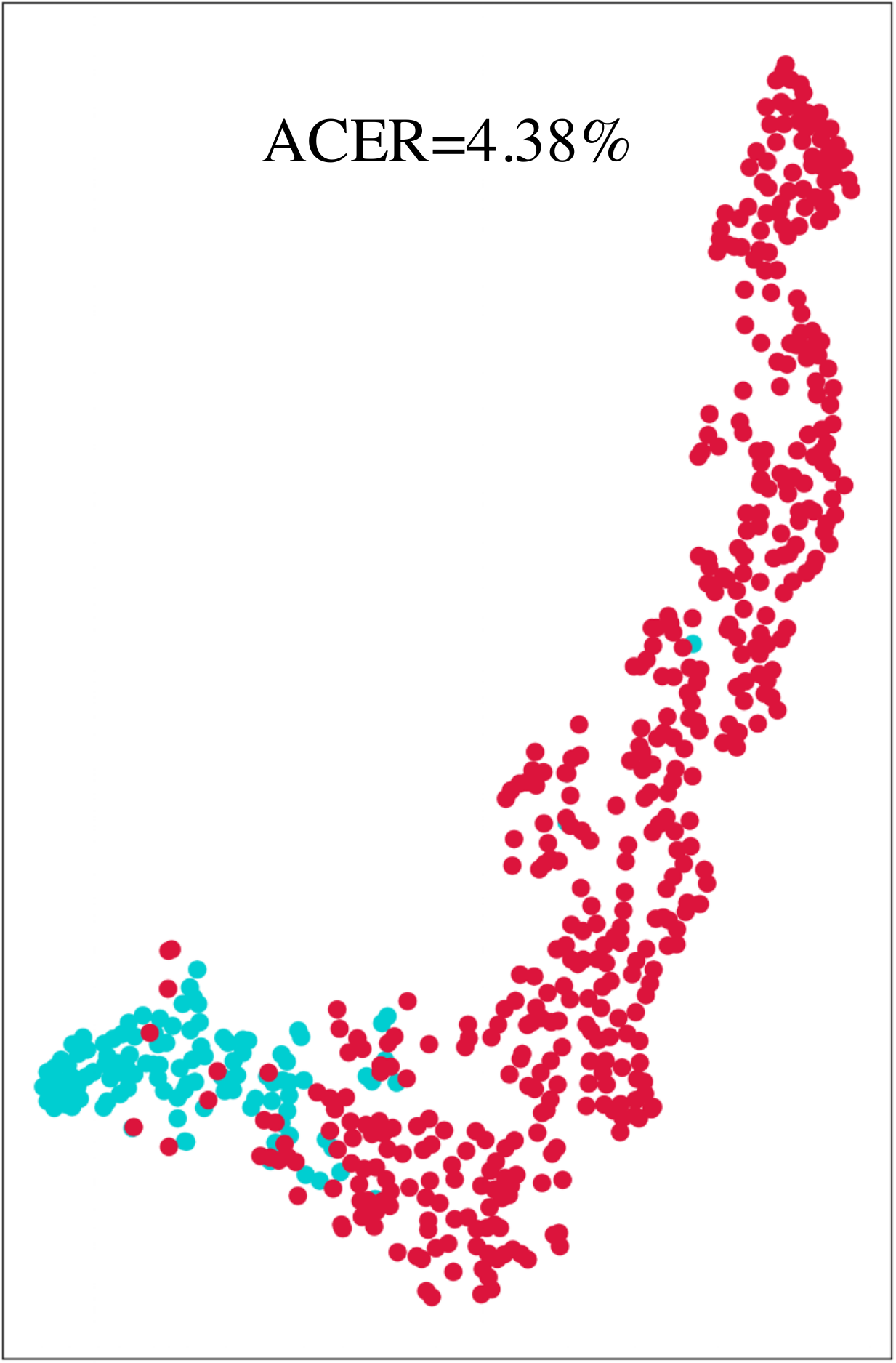}
    \caption{}
    \label{fig:tsne_a} 
  \end{subfigure}
  \begin{subfigure}[b]{0.141\textwidth}
    \includegraphics[width=1.0\linewidth]{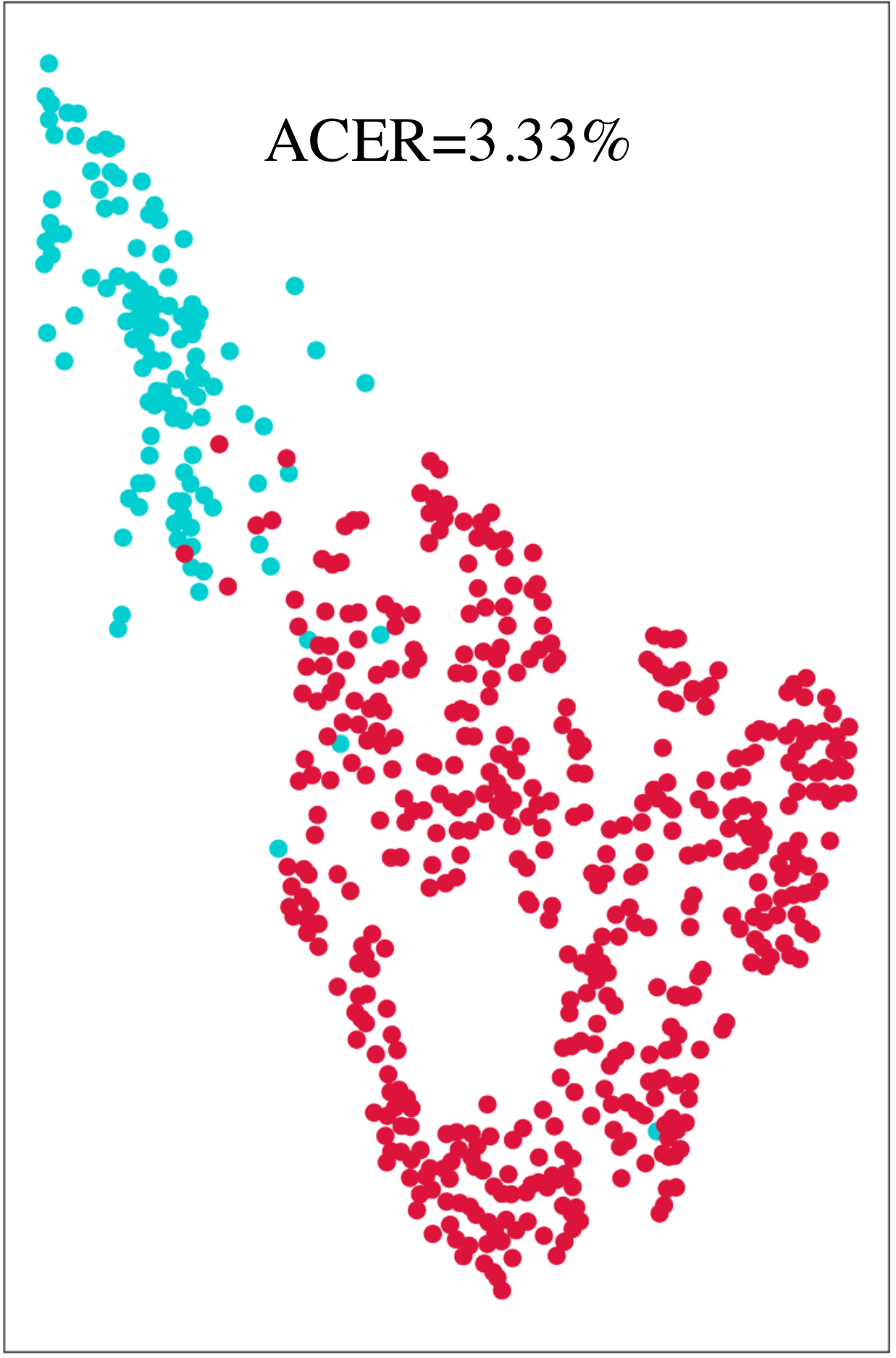}
    \caption{}
    \label{fig:tsne_b} 
  \end{subfigure}
  \begin{subfigure}[b]{0.141\textwidth}
    \includegraphics[width=1.0\linewidth]{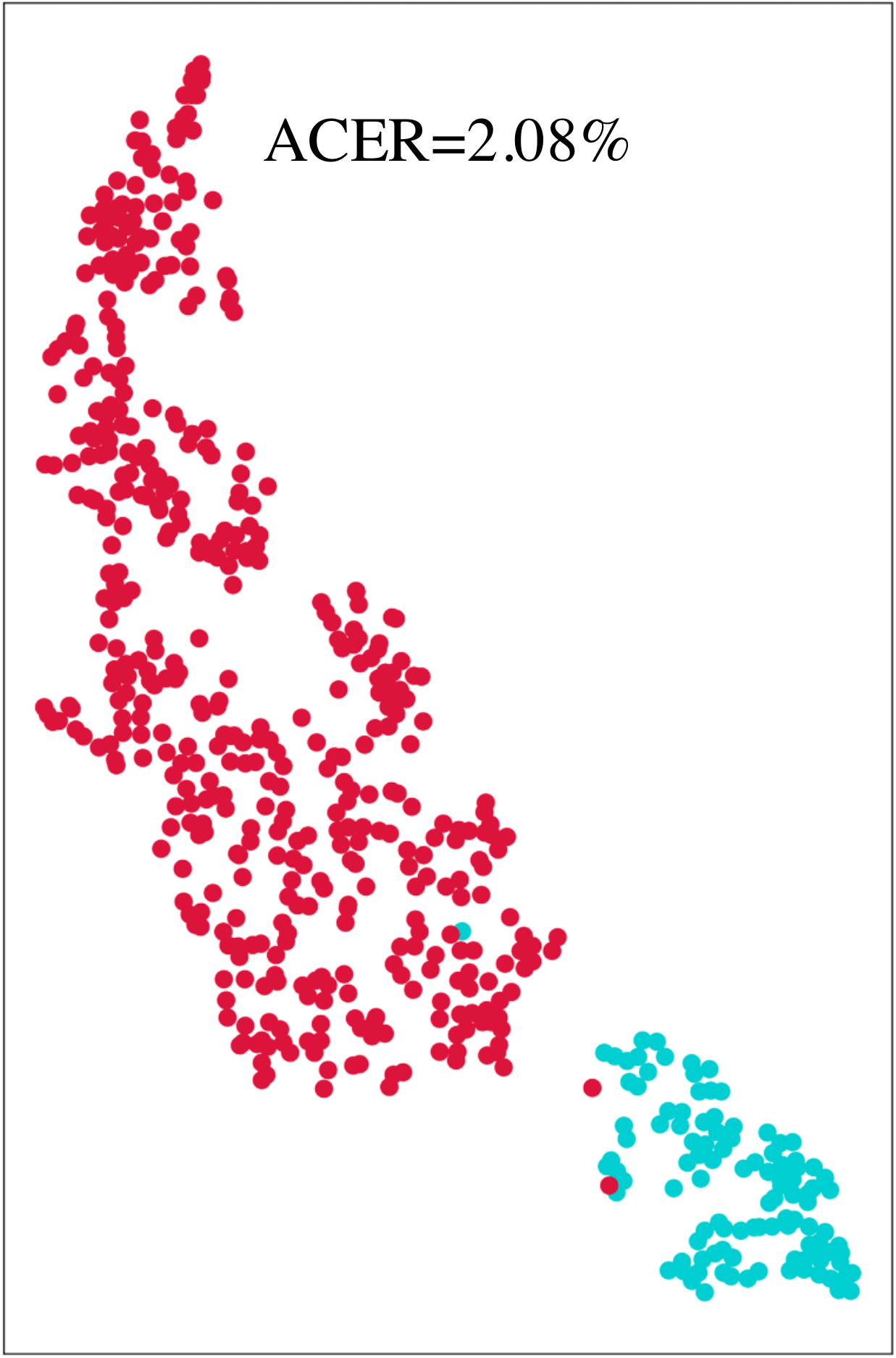}
    \caption{}
    \label{fig:tsne_c} 
  \end{subfigure}
  \caption{Feature distribution of the testing samples on \emph{SSL-protocol-1} using t-SNE~\cite{maaten2008visualizing}. Left: CDCN~\cite{yu2020searching} baseline with 20\% labeled data, Middle: EPCR supervised only with 20\% labeled data, Right: EPCR  with 20\% labeled data and the remaining 80\% unlabeled data.}
  \label{fig:tsne} 
  \vspace{-0.8em}
\end{figure}

The consistency regularization~\cite{chen2020simple, tarvainen2017mean} usually performs upon reasonable representation perturbation. Thus, it is crucial to design suitable augmentations to generate useful perturbations that can enhance the model's output invariance. Besides the common augmentations utilized in both FAS and SSL (i.e., random crop, random light/contrast, random erasing, and cutout~\cite{devries2017improved,chen2020simple}), we introduce an extra augmentation method named Patch Shuffle Augmentation (PSA) specially for FAS, which has been proposed in ~\cite{zhang2021structure}. 
As shown in Fig.~\ref{fig:spa}, we evenly divide the face input into $K$ patches. Then we shuffle these patches and splice them back into a piece of the original size. The design philosophy of PSA is according to the assumption that every patch of the face image may contain similar live/spoof clues. PSA would bring diverse spatial and  structural perturbation without breaking intrinsic live/spoof patterns. $K$ is fixed on $9$ in our experiments.

As shown in Tab.~\ref{tab:ablation_augmentation}, we conduct experiments on EPCR with different combinations of augmentation methods. From ``Crop'' to ``Color'' to ``Flip'' to ``Cutout'', the performance is getting better as the increase of augmentation methods, which implies that these four augmentations bring benefit to the FAS method. When ``Blur'' is added as the fifth augmentation method, we find that the ACER becomes higher than that of the previous combination. This indicates that the Gaussian Blur will disturb the spoof cues and may be not suitable for the FAS method. Moreover, we study the effect of patch number in Patch Shuffle Augmentation (PSA). Tab.~\ref{tab:ablation_augmentation} shows that $K=9 (3 \times 3)$ is the best choice for PSA, and PSA indeed improves the performance of FAS method. According to the above-mentioned study, we utilize ``Crop'', ``Color'',  ``Flip'', ``Cutout'' and ``PSA $3 \times 3$'' as our augmentation methods.

\section*{F. Feature Distribution Visualization}
As shown in Fig.~\ref{fig:tsne}, we visualize the feature distribution of the testing samples on \emph{SSL-protocol-1} using t-SNE~\cite{maaten2008visualizing}. We can see that samples in Fig.~\ref{fig:tsne_c} are the best-clustered, and Fig.~\ref{fig:tsne_b} presents better-clustered behavior than that of Fig.~\ref{fig:tsne_a}. Fig.~\ref{fig:tsne} demonstrates that EPCR outperforms the CDCN baseline, and introducing unlabeled facial images can improve the the performance of the FAS model.

\end{document}